%% file: main.tex
\documentclass[11pt,a4paper]{article}

\usepackage[T1]{fontenc}
\usepackage[utf8]{inputenc}
\usepackage{lmodern}
\usepackage{microtype}
\usepackage[a4paper,left=22mm,right=22mm,top=30mm,bottom=30mm,
  headheight=40pt,headsep=19pt,footskip=30pt]{geometry}
\usepackage{xcolor}
\definecolor{reportnavy}{HTML}{183451}
\definecolor{reportciteblue}{HTML}{337AB7} 
\definecolor{reportrule}{gray}{0.5}
\usepackage{graphicx}
\usepackage{fancyhdr}
\usepackage{titlesec}
\usepackage{amsmath,amssymb,bm}
\usepackage{booktabs,multirow,tabularx,makecell,array}
\usepackage{xspace,pifont,adjustbox,placeins}
\usepackage{float,needspace}
\usepackage{wrapfig}
\usepackage{ragged2e}
\newcolumntype{Y}{>{\RaggedRight\arraybackslash}X}
\usepackage{caption,subcaption}
\DeclareCaptionFont{teaserfont}{\fontsize{8.5}{10.2}\selectfont}
\usepackage{algorithm}
\usepackage{algpseudocode}
\usepackage{enumitem}
\usepackage[numbers,square,sort&compress]{natbib}
\usepackage[
    unicode,
    colorlinks=true,
    allcolors=.,
    citecolor=reportciteblue
]{hyperref}
\usepackage[capitalise,noabbrev]{cleveref}

\renewcommand{\thesection}{\arabic{section}}
\titleformat{\section}{\normalfont\fontsize{13}{16}\selectfont\bfseries\color{reportnavy}}{\thesection.}{0.55em}{}
\titleformat{\subsection}{\normalfont\normalsize\bfseries}{\thesubsection.}{0.55em}{}
\titleformat{\subsubsection}[runin]{\normalfont\normalsize\bfseries}{\thesubsubsection.}{0.55em}{}[.]
\titlespacing*{\section}{0pt}{1.05\baselineskip}{0.35\baselineskip}
\titlespacing*{\subsection}{0pt}{0.65\baselineskip}{0.20\baselineskip}
\titlespacing*{\subsubsection}{0pt}{0.65\baselineskip}{0.5em}
\setlist{nosep,leftmargin=*}

\newcommand{\PaperTitle}{ScoutNav: Look--Act Decoupling for Cost-Efficient\\Agentic Vision-and-Language Navigation}
\newcommand{\RunningTitle}{ScoutNav: Look--Act Decoupling for Cost-Efficient Agentic Vision-and-Language Navigation}
\newcommand{\PaperAuthors}{%
  \mbox{Bin Wang\textsuperscript{1,4,5}},
  \mbox{Zihao Xin\textsuperscript{2}},
  \mbox{Tuo Chen\textsuperscript{2}},
  \mbox{Weihao Luo\textsuperscript{3}},
\mbox{Lianyu Hu\textsuperscript{5}},\\
  \mbox{Runmin Cong\textsuperscript{4}},
  \mbox{Wentong Li\textsuperscript{2, \dag}},
  \mbox{Wei Zhang\textsuperscript{4, \dag}},
  \mbox{Xudong Jiang\textsuperscript{5}}}
\newcommand{\PaperAffiliations}{%
  \textsuperscript{1}Shandong University of Technology,
  \textsuperscript{2}Nanjing University of Aeronautics and Astronautics,
  \textsuperscript{3}Ningbo University,
  \textsuperscript{4}Shandong University,
  \textsuperscript{5}Nanyang Technological University
  }
\newcommand{\PaperNotes}{%
  \textsuperscript{\dag}Corresponding authors}

\newcommand{\ProjectURL}{https://github.com/BBYL9413/ScoutNav}
\newcommand{\SDUTLogoFile}{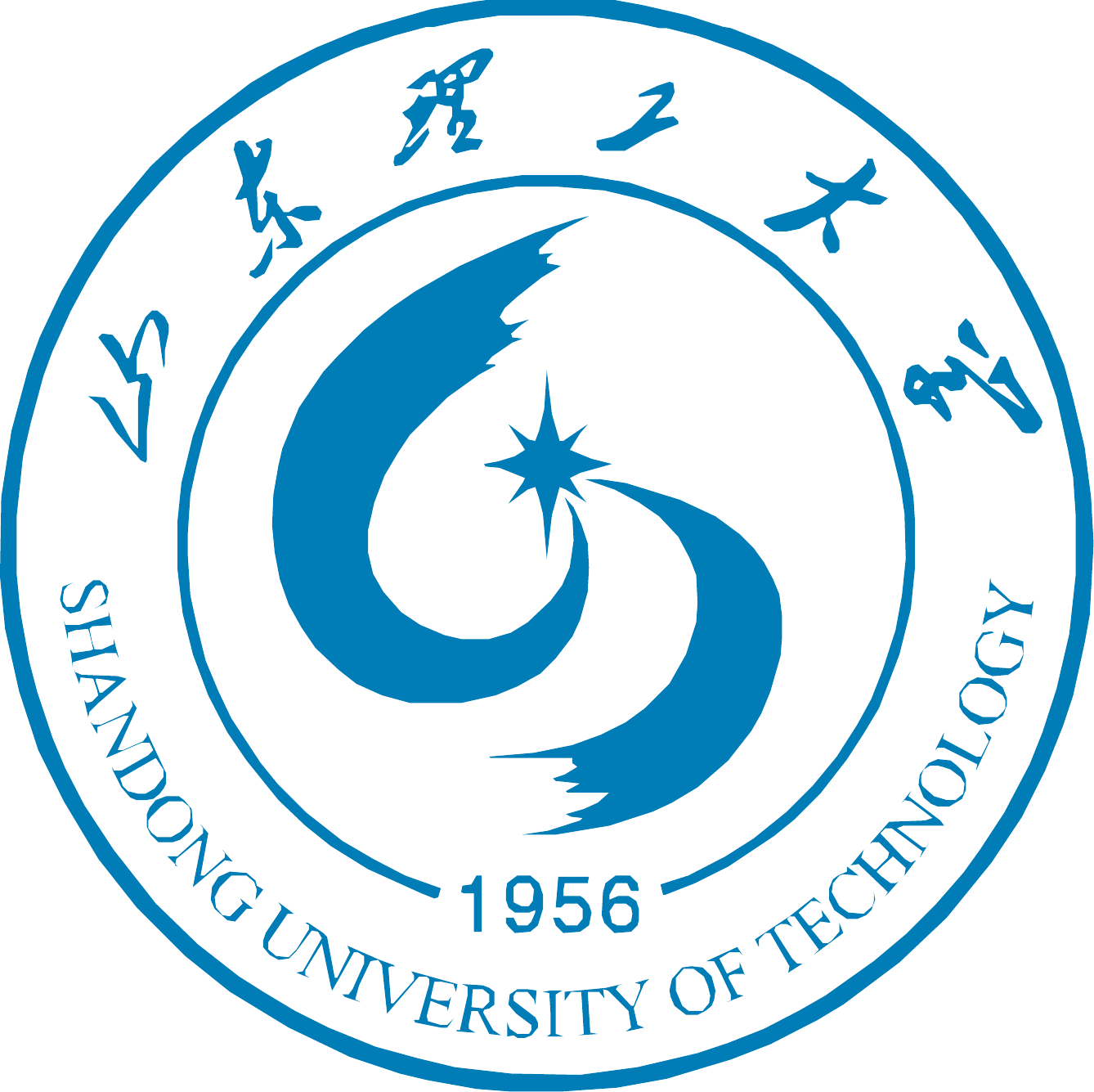}
\newcommand{\SDULogoFile}{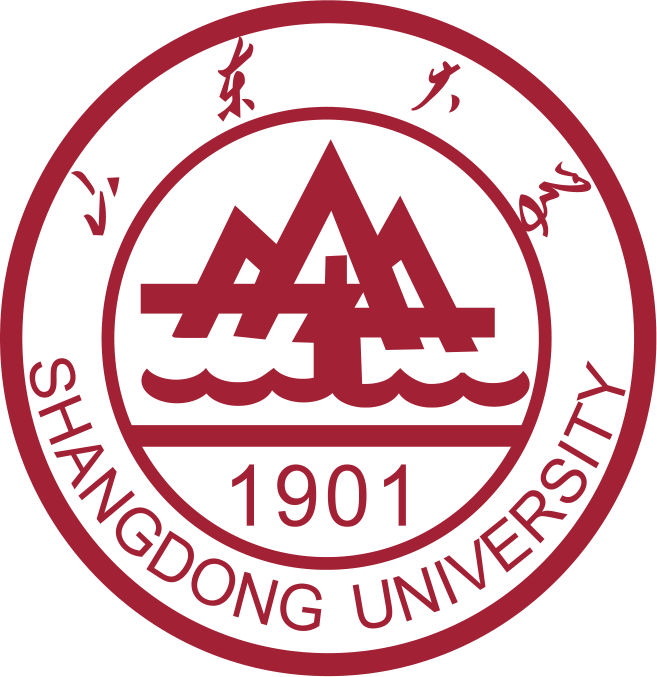}
\newcommand{\NBULogoFile}{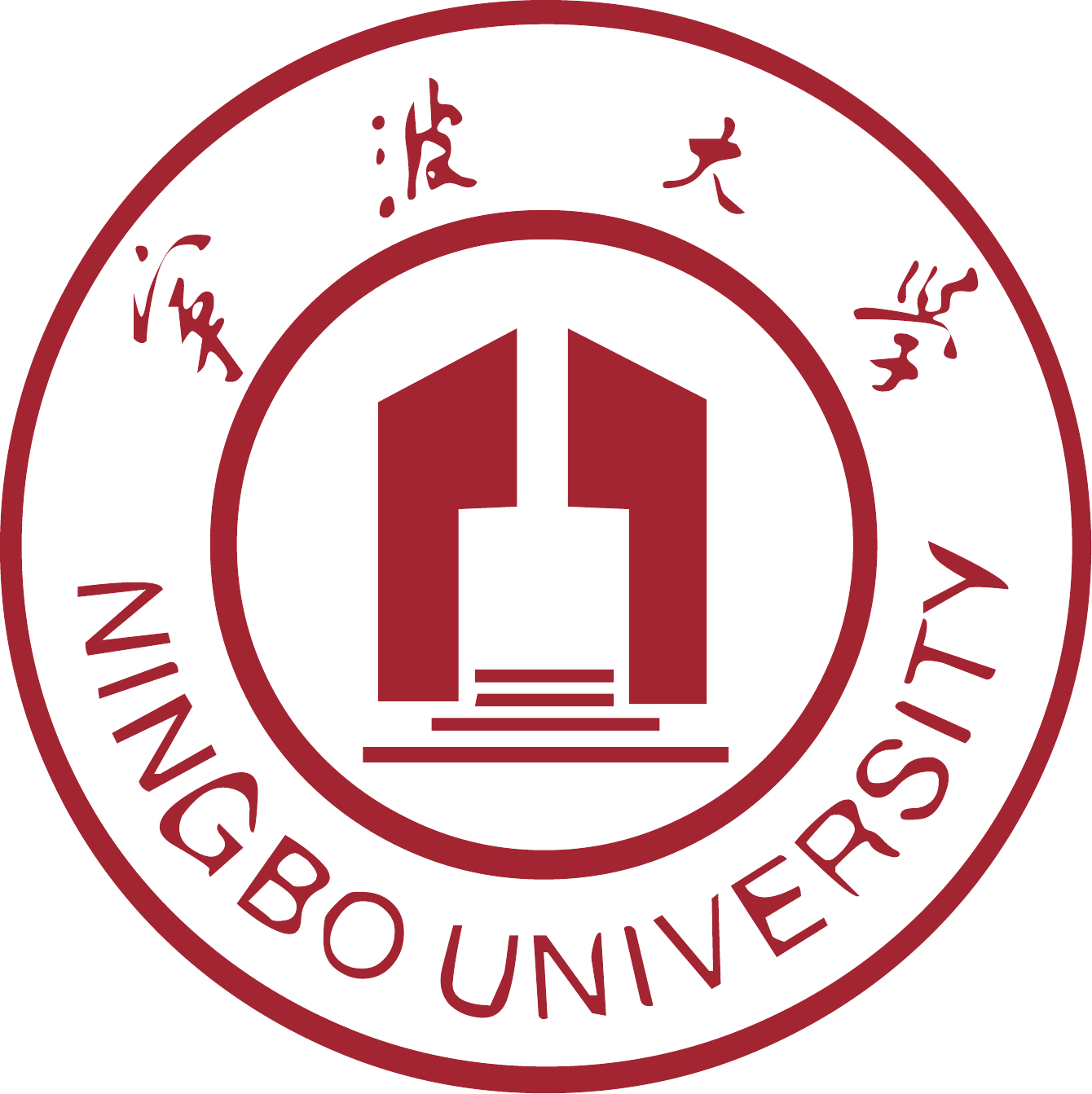}
\newcommand{\NUAALogoFile}{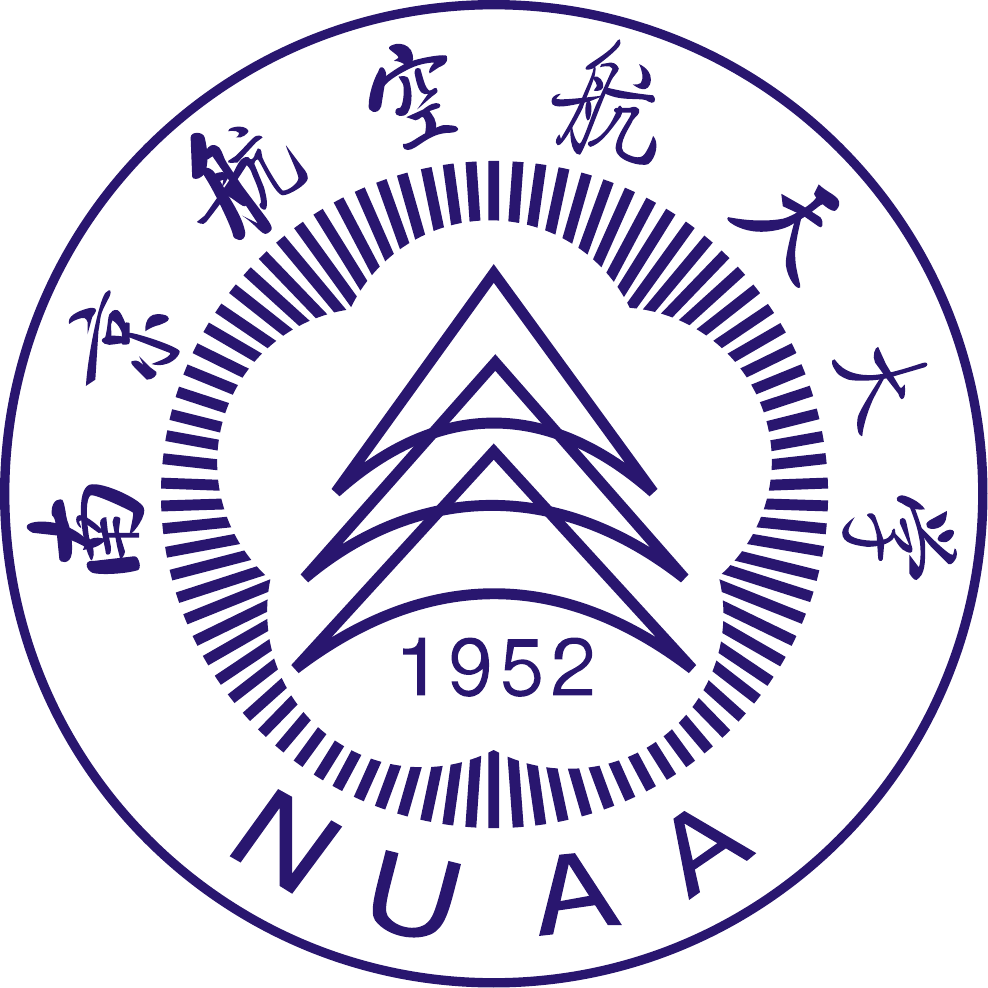}
\newcommand{\NTULogoFile}{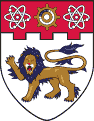}
\newcommand{\HeaderLogoHeight}{33pt}
\newcommand{\FirstPageFooter}{}

\newcommand{\HeaderEmblem}[2]{%
  \IfFileExists{#1}{\raisebox{-0.5\height}{\includegraphics[height=\HeaderLogoHeight,keepaspectratio]{#1}}}%
  {\raisebox{-0.5\height}{\sffamily\bfseries\color{reportnavy}#2}}}
\newcommand{\HeaderLogoGap}{10pt}
\newcommand{\ReportLogoRow}{%
  \HeaderEmblem{\SDUTLogoFile}{SDUT}\hspace{\HeaderLogoGap}%
  \HeaderEmblem{\SDULogoFile}{SDU}\hspace{\HeaderLogoGap}%
  \HeaderEmblem{\NBULogoFile}{NBU}\hspace{\HeaderLogoGap}%
  \HeaderEmblem{\NUAALogoFile}{NUAA}\hspace{\HeaderLogoGap}%
  \HeaderEmblem{\NTULogoFile}{NTU}}

\fancypagestyle{firstpage}{%
  \fancyhf{}%
  \fancyhead[R]{\ReportLogoRow\rule[-\dimexpr\HeaderLogoHeight/2+3pt\relax]{0pt}{0pt}}%
  \fancyfoot[L]{\footnotesize\FirstPageFooter}%
}

\makeatletter
\renewcommand{\maketitle}{%
  \thispagestyle{firstpage}%
  \noindent{\color{reportnavy}\bfseries\fontsize{18}{22}\selectfont\PaperTitle\par}%
  \vspace{15pt}%
  \noindent{\small\bfseries\PaperAuthors\par}%
  \vspace{-5pt}%
  \noindent{\fontsize{10}{12}\selectfont\PaperAffiliations\par}%
  \vspace{-6pt}%
  \noindent{\fontsize{10}{12}\selectfont\PaperNotes\par}%
  \vspace{-2pt}%
  \ifx\ProjectURL\empty\else
    \noindent{\fontsize{10}{12}\selectfont\color{reportnavy}\expandafter\url\expandafter{\ProjectURL}\par}%
  \fi
  \vspace{13pt}%
}
\renewenvironment{abstract}{%
  \par\noindent\begingroup\fontsize{10}{13.2}\selectfont
  \setlength{\parindent}{0pt}\setlength{\parskip}{0pt}%
  \textbf{Abstract:}\par\noindent\ignorespaces
}{\par\endgroup\vspace{0.7\baselineskip}}
\makeatother
\hypersetup{pdftitle={ScoutNav: Look--Act Decoupling for Cost-Efficient Agentic Vision-and-Language Navigation}}
\input{preamble}

\DeclareUnicodeCharacter{2011}{\mbox{-}}

\begin{document}
\maketitle
\input{sec/0_abstract}
\input{sec/1_intro}
\input{sec/2_related}
\input{sec/3_method}
\input{sec/4_experiments}
\input{sec/5_conclusion}

\FloatBarrier

{\small
\setlength{\bibsep}{0.22\baselineskip}
\bibliographystyle{unsrtnat}
\bibliography{references}
}

\appendix
\input{sec/X_suppl}

\end{document}

%% file: preamble.tex
\makeatletter
\DeclareRobustCommand\onedot{\futurelet\@let@token\@onedot}
\def\@onedot{\ifx\@let@token.\else.\null\fi\xspace}
\makeatother
\def\eg{\emph{e.g}\onedot} 
\def\ie{\emph{i.e}\onedot} 
 
 \def\vs{\emph{vs}\onedot}

\newcommand{\method}{ScoutNav\xspace}        
\newcommand{\explorer}{Scout\xspace}          
\newcommand{\decider}{Decider\xspace}
\newcommand{\packet}{evidence packet\xspace}
\newcommand{\sketch}{scene sketch\xspace}

\newcommand{\tool}[1]{\texttt{#1}}
\newcommand{\qd}{\tool{query\_depth}\xspace}
\newcommand{\rc}{\tool{recall}\xspace}
\newcommand{\mt}{\tool{move\_to}\xspace}
\newcommand{\st}{\tool{stop}\xspace}

\newcommand{\Tinfo}{\mathcal{T}_{\text{info}}}
\newcommand{\Tact}{\mathcal{T}_{\text{act}}}

%% file: sec/0_abstract.tex
\begin{abstract}
Tool-calling vision-language models (VLMs) enable agentic navigation by acquiring spatial evidence and retrieving visual memories before acting.
However, invoking a powerful API VLM model throughout this process makes each movement depend on multiple costly inference rounds.
We introduce \method, an agentic framework for cost-efficient vision-and-language navigation in continuous environments based on \emph{look--act decoupling}.
The key idea is to allocate model capacity according to an operation's effect on the trajectory: gather evidence locally and retain the stronger model for action selection.
A local \explorer queries depth and recalls past observations until it proposes moving or stopping, at which point control passes to a core VLM \decider without executing the proposal.
The \decider receives current observations and a compact \packet augmented with a geometric \sketch, and can request further evidence before choosing an action.
When a waypoint fails the geometric feasibility check, bounded local repair searches for an executable target within the \decider's intended direction.
This division of work requires neither navigation-specific training nor a learned routing model.
On the Rand100 protocol for R2R-CE val-unseen, ScoutNav achieves a success rate of up to 86.0\% while reducing costs by nearly half.
\end{abstract}

%% file: sec/1_intro.tex
\vspace{-20pt}
\section{Introduction}
\label{sec:intro}

Vision-and-language navigation (VLN) asks an embodied agent to follow natural-language directions through an unfamiliar environment~\citep{anderson2018vision,xin2026decovln,xininstruction,jiang2026geoscaffold}.
In continuous environments, the agent must ground landmarks in its observations, track progress along the instruction, and translate a chosen route into feasible movement~\citep{krantz2020beyond}.
Pretrained language and vision-language models offer a way to perform these tasks without navigation-specific policy training~\citep{zhou2024navgpt,qiao2025open}.
Recent advances improve how these models use spatial structure, execution feedback, and navigation history~\citep{dai2026history, he2026strider}, bringing the interface between model reasoning and embodied interaction into focus.

Tool-calling navigation agents make this interface explicit.
AgenticNav~\citep{li2026agenticnav} lets a VLM query metric depth, retrieve earlier views, and select pixel targets that are converted into executable motion.
HarnessVLN~\citep{chen2026harnessvln} further emphasizes coordination between planning, spatial evidence, memory, and execution feedback through an agent harness.
These interfaces allow an agent to seek information as a decision unfolds.
Yet when one powerful VLM model controls the entire tool loop, every intermediate query incurs another inference round over visual observations and an expanding interaction history.
Therefore, to make this navigation approach practical, it is necessary to reduce reliance on costly model invocations while maintaining its ability to gather evidence.

Prior work on efficient VLN lowers the cost of each inference, through compact navigation architectures such as COSMO~\citep{zhang2025cosmo} or through selective perception and context selection as in ProFocus~\citep{xue2026profocus}, but still routes every decision through a single model.
In a tool-calling agent, a decision instead consists of several model calls, which need not be served by the same model.

\newpage
\setlength{\intextsep}{2pt}
\begin{wrapfigure}{r}{0.5\textwidth}
  \captionsetup{font=teaserfont,skip=3pt}
  \centering
  \includegraphics[width=\linewidth]{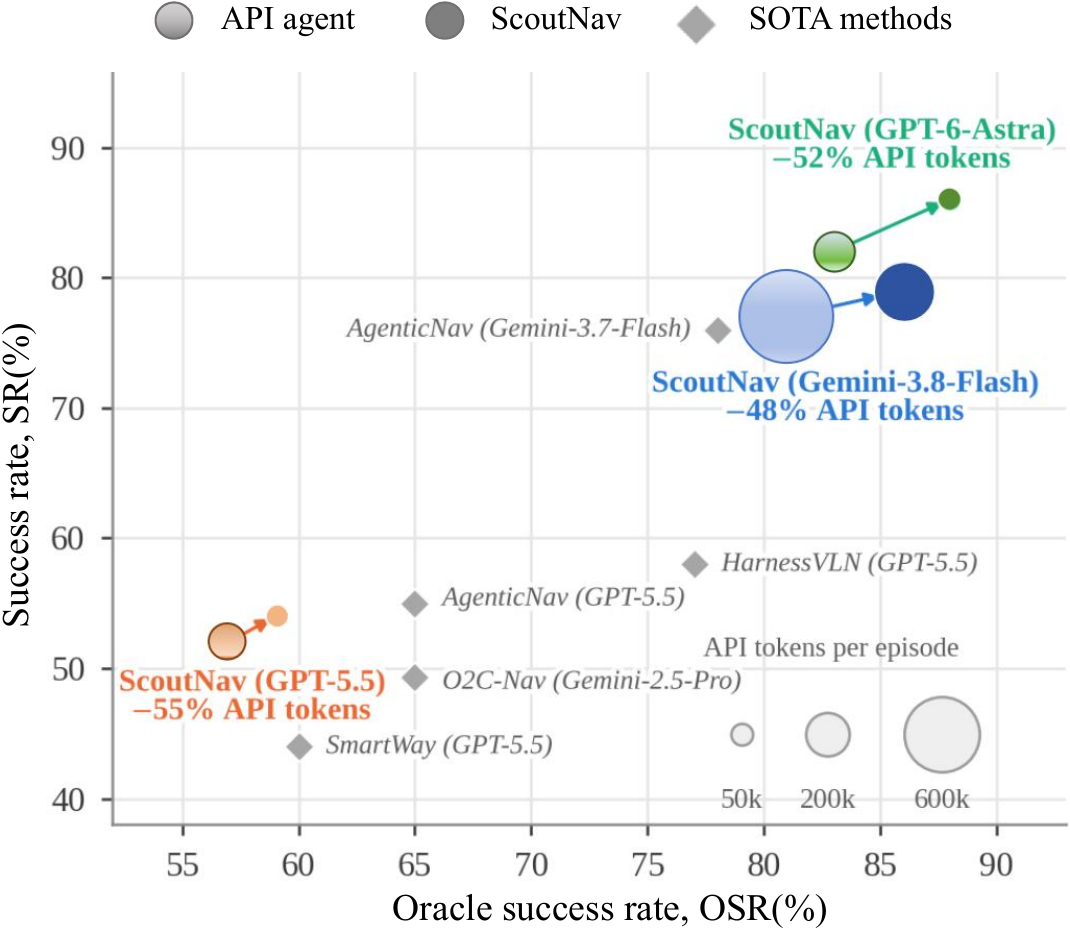}
  \vspace{1mm}
  \caption{\textbf{Success rate \vs Oracle success rate
  on R2R-CE val-unseen (Rand100).}
  \method delegates evidence gathering to a local \explorer and hands action selection to a \decider, which may request additional evidence.
  With Gemini-3.8-Flash, GPT-5.5 and GPT-6-Astra as the API model, it matches the success rate of our API-only AgenticNav baseline with 48\%, 55\% and 52\% fewer API tokens per episode (input, output, and reasoning).
  Token counts follow each provider's accounting, all points are single runs on the same 100 episodes.}
  \label{fig:teaser}
\end{wrapfigure}

We study a complementary question: \textit{How should a local model and an API model share the work within a navigation decision?}
Model cascades~\citep{chen2023frugalgpt,ong2025routellm} answer a related question by routing each query as a whole to one model.
Applied to navigation, however, such routing assigns both the evidence gathering of a step and the resulting action to the same model, although the two carry different stakes.
It also requires predicting, from partial observations, when the local model's action can be trusted.
In contrast, the tool interface provides a signal that requires no such prediction: whether the next operation gathers information or changes the agent's trajectory.

This distinction motivates \emph{look--act decoupling}.
A depth query or a recall of a previous view can be followed by further inspection without moving the agent; choosing a waypoint changes its position and subsequent observations, while stopping ends the episode.
Evidence acquisition can still omit or misinterpret an important cue, so its output should remain available for a stronger model to examine before committing to motion.
The opportunity is to delegate the preparatory tool loop locally while retaining the stronger model's judgment for route selection and termination.
In summary, this division also targets the main source of expense in tool-calling navigation agents, which spend most of their interaction on gathering evidence rather than on acting \citep{li2026agenticnav, chen2026harnessvln, dai2026gpt}.

\setlength{\intextsep}{12pt plus 2pt minus 2pt}

We introduce \method, an agentic framework that implements this division of work for tool-calling navigation agents (\cref{fig:pipeline}).
At each decision step, a local open-weight VLM, the \explorer, gathers depth measurements and recalls relevant observations.
Its first proposal to move or stop triggers a handoff to an API VLM, the \decider; the proposal is discarded, and the \decider selects the action itself.
To make this handoff economical, we assemble a compact \packet containing the current observations and retrieved evidence, augmented with a geometric \sketch computed from depth.
The packet transfers the results of exploration without replaying the \explorer's full conversation, and the \decider may request additional evidence before acting.
Finally, when a selected waypoint fails the geometric feasibility check, the \explorer searches for a feasible target within a bounded deviation from the \decider's intended heading.
This local repair handles geometric adjustments while preserving the selected direction; unsuccessful repair returns control to the \decider.
The framework requires no navigation-specific training or confidence calibration.

We evaluate \method on the fixed Rand100 protocol of R2R-CE val-unseen~\citep{qiao2025open,shi2025smartway,li2026agenticnav,xin2026agentvln}, using the same observation and action interface as our API-only baseline.
Across API models, \method achieves a higher success rate than the API-only agent while lowering the API cost by 45--50\% and the number of API tokens by 48--55\% (\cref{fig:teaser}); with a stronger API model, it reaches 86.0\% success.

\Needspace{5\baselineskip}
Our contributions are threefold:
\begin{itemize}
  \item We introduce look--act decoupling for tool-calling agentic VLN task, assigning local evidence gathering and API action selection according to the proposed operation's effect on the trajectory. Use a well-designed harness to decouple tasks and reduce API costs.
  \item We develop \method with a compact \packet, a geometric \sketch, and bounded waypoint repair to support this collaboration without training a navigation policy or a router.
  \item We show on R2R-CE Rand100 that \method lowers the API cost by nearly half across API models while achieving a higher success rate, and find that at handoff the API model lacks visual context rather than deliberation, which indicates what future collaboration between local and API models should transfer.
\end{itemize}

%% file: sec/2_related.tex
\section{Related Work}
\label{sec:related}

\textbf{Agentic vision-and-language navigation.}
Agentic VLN use pretrained language and vision-language models to interpret instructions and guide navigation without task-specific policy training~\citep{zhou2024navgpt,qiao2025open}.
Existing approaches improve navigation through structured spatial reasoning and action generation~\citep{chen2024mapgpt,chen2025constraint,he2026strider, shi2025smartway}, or through the organization of visual history and prior experience~\citep{dai2026history, deng2026spacevln, liu2026ham}.
These directions address complementary requirements: grounding instructions in navigable space and maintaining context across decisions.
Recent agent harnesses bring perception, memory, and execution into an iterative tool-calling process~\citep{li2026agenticnav,chen2026harnessvln,zhou2026embodied}, enabling models to acquire evidence and respond to execution feedback during navigation.
We build on AgenticNav's pixel-based action interface and depth and recall tools.
Our focus is how local and API models share this interaction: the local model gathers initial evidence, while the API model selects an action and can request further information.

\noindent\textbf{Model cascades and cost-efficient agents.}
Efficiency in navigation has been pursued through compact models, selective processing of visual context, and fewer inference rounds~\citep{zhang2025cosmo,xue2026profocus, pan2026one}.
A complementary direction distributes planning, grounding, and execution across models or modules of different capacity~\citep{ding2026lavira, ding2026uni, wang2025clash, xiong2026sfco}.
More generally, cascades and adaptive agents allocate computation by choosing a model for each query or step~\citep{chen2023frugalgpt, piatrashyn2026redact, younesi2026need, ong2025routellm}, or adjusting its reasoning budget~\citep{lee2026dart, yang2026ares}.
These approaches motivate sharing work across models while retaining stronger reasoning where needed.
Trajectory-level speculation also uses the boundary between read-only and state-changing tools: a weak model executes information queries, and a strong model verifies the drafted segment before committing an action~\citep{wang2026trajectory}.
\method uses this boundary to hand over action generation.
The local draft is discarded, and the API \decider selects its own action from current observations and a compact \packet, without receiving the full local conversation.
The handoff requires no confidence estimate or learned router, and bounded local repair handles geometric adjustments to the selected waypoint.

%% file: sec/3_method.tex
\section{Method}
\label{sec:method}

\begin{figure}[!htbp]
  \centering
  \includegraphics[width=\linewidth]{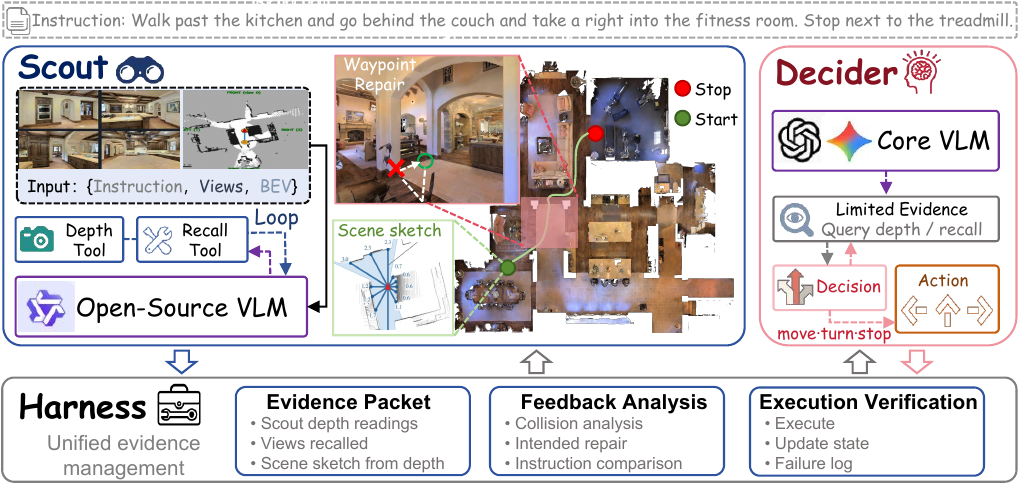}
  \caption{\textbf{Overview of \method.}
  The \explorer looks, the \decider acts, and the Harness ties them together.
A local \explorer gathers evidence for free; the Harness distills it into a compact \packet with a \sketch, hands the decision to an API \decider, verifies every move, and repairs rejected waypoints without changing their intent.
Without any training or learned router, this harness roughly halves the API cost of a tool-calling agent without losing success}
  \label{fig:pipeline}
\end{figure}

\textbf{Observation and tools.}
We follow the AgenticNav harness~\citep{li2026agenticnav}.
At decision step $t$ the agent receives the instruction $L$, four RGB-D views $\{I^k_t, D^k_t\}_{k=1}^{4}$ and the BEV occupancy map $M_t$ built from the depth point clouds, and a textual history $H_t$ containing the reasons and executed actions of the last six steps.
The VLM interacts with the environment through four tools, which we split into
\begin{equation}
  \begin{aligned}
    \Tinfo &= \{\qd,\ \rc\},\\
    \Tact &= \{\mt,\ \st\}.
  \end{aligned}
\end{equation}
$\qd(k,u,v)$ returns metric depth statistics at normalized pixels of view $k$, and $\rc(p,k)$ returns view $k$ observed at a past decision point $p$.
$\mt(k,u,v)$ back-projects a pixel to a 3D target, rotates the agent towards it, and moves straight for $\rho=\min(\rho_{\max}, \|x\|-m)$ meters; the motion is executed only if a corridor of half-width $r_a$ along the heading is free of obstacle points, and otherwise the tool returns \emph{reselect}.
A step ends with an accepted \mt or with \st, and each step has a budget of tool rounds.
Calls in $\Tinfo$ are read-only, whereas calls in $\Tact$ change the agent pose or terminate the episode.

\textbf{Cost model.}
For an API model with prices $c_{\text{in}}$ and $c_{\text{out}}$ per token, the cost of an episode is
\begin{equation}
  C = \sum_{j \in \mathcal{A}} \big( c_{\text{in}}\, n^{\text{in}}_j + c_{\text{out}}\, (n^{\text{out}}_j + n^{\text{think}}_j) \big),
\end{equation}
where $\mathcal{A}$ is the set of API calls and $n^{\text{in}}_j$, $n^{\text{out}}_j$, $n^{\text{think}}_j$ are the input, output, and reasoning tokens of call $j$.
Because each tool round re-sends the step context, \ie, four views, the BEV map, and all previous rounds of the step, the input term dominates.
In the API-only VLM agent, rounds that end in \qd or \rc account for a large part of $C$.
Calls to the local VLM model incur no API cost; we report their number and tokens separately.

\subsection{Look--act decoupling with intent-triggered handoff}
\label{sec:gate}

We refer to calls in $\Tinfo$ as \emph{looking} and to calls in $\Tact$ as \emph{acting}; \method decouples the two and assigns them to two models (\cref{fig:pipeline}, \cref{alg:step}).
The \explorer $E$ is a local open-weight VLM model queried without extended reasoning; the \decider $D$ is an API VLM model queried with its default reasoning setting.

Each decision step starts with the \explorer, which receives the same step context $X_t$ as the original agent and issues tool calls round by round.
Let $a$ denote the calls of the \explorer in the current round.
If $a \subseteq \Tinfo$, the calls are executed and their results are appended to the \explorer's context.
Otherwise, \ie, as soon as the \explorer proposes to move or stop, the step is \emph{escalated}: the proposed call is neither executed nor shown to the \decider, and the \decider takes over the remainder of the step.
The draft thus serves only as a trigger: the \explorer determines \emph{when} the evidence suffices for a decision, and the \decider determines \emph{which} action to take.
A step also escalates when the \explorer exhausts its round budget, so every decision step escalates exactly once.
The \decider may request up to $K$ further information calls before it must act; in this forced round only \mt is offered.
We do not offer \st in a forced round because an API model that must choose between moving and stopping before it is ready tends to stop prematurely.

The handoff requires neither a confidence estimate nor a learned router.
It exploits the asymmetry between the two tool sets: an erroneous information request of the \explorer costs a local tool round but leaves the agent state unchanged, whereas every state change is selected by the \decider or, after a rejected waypoint, constrained to its intended heading.
Information gathering, which accounts for most API calls of the original agent, is thereby moved to the local model.

\begin{algorithm}[!htbp]
\caption{One decision step of \method.}
\label{alg:step}
\small
\begin{algorithmic}[1]
\Require step context $X_t=(L, I_t, M_t, H_t)$, \explorer $E$, \decider $D$, budget $K$
\State $\mathcal{E} \gets \emptyset$ \Comment{\explorer evidence}
\Loop
  \State $a \gets E(X_t, \mathcal{E})$ \Comment{local, no API cost}
  \If{$a \not\subseteq \Tinfo$} \textbf{break} \Comment{escalate; discard $a$}
  \EndIf
  \State $\mathcal{E} \gets \mathcal{E} \cup \textsc{Execute}(a)$
\EndLoop
\State $P_t \gets \textsc{Packet}(X_t, \mathcal{E}, \textsc{Sketch}(D_t))$ \Comment{\cref{sec:packet}}
\For{$i = 0, 1, \dots$}
  \State $b \gets D(P_t;\ \text{tools} = \Tinfo \cup \Tact \text{ if } i < K \text{ else } \{\mt\})$ \Comment{API}
  \If{$b = \st$} \Return \st
  \ElsIf{$b = \mt(\cdot)$ is executable} \Return $b$
  \ElsIf{$b = \mt(\cdot)$ is rejected}
    \State $b' \gets \textsc{Repair}_E(b)$ \Comment{$\varnothing$ if no executable move; \cref{sec:repair}}
    \If{$b' \neq \varnothing$} \Return $b'$ \EndIf
  \EndIf
  \State $P_t \gets P_t \cup \textsc{Execute}(b)$
\EndFor
\end{algorithmic}

\end{algorithm}
\subsection{Evidence packet and scene sketch}
\label{sec:packet}

\textbf{Evidence packet.}
Passing the \explorer's tool conversation to the \decider would reintroduce the cost we aim to remove and expose the \decider to the \explorer's intermediate reasoning.
Instead, the \decider receives an \packet $P_t$ that contains
(i)~the step observation, \ie, the four views, the BEV map, and the textual history, given once;
(ii)~the depth readings gathered by the \explorer, rendered as text;
(iii)~the past views recalled by the \explorer; and
(iv)~a geometric \sketch computed by the harness from the depth point cloud.

\textbf{Scene sketch.}
The \sketch is derived from the \decider's own information requests.
When the packet contained only items (i)--(iii), about half of the escalated steps spent the entire read budget on \qd, and most of the queried pixels lay on the floor; the \decider mostly asked where the floor is and how far it can walk.
Both questions can be answered exactly from depth, so the harness computes the answers in advance and serializes them as text.
The \emph{floor depth grid} reports the depth at a $3\times3$ grid of pixels $u \in \{0.25, 0.5, 0.75\}$, $v \in \{0.65, 0.80, 0.95\}$ in each view, which covers the floor region where waypoints are selected.
The \emph{free-space profile} reports, for headings $\theta \in \{-150^\circ, -120^\circ, \dots, 180^\circ\}$, the distance to the nearest obstacle point inside a corridor of half-width $r_a$ along $\theta$, capped at 6\,m:
\begin{equation}
  f_t(\theta) = \min\Big(6,\ \min_{x \in \mathcal{O}_t} \{\langle x, \mathbf{b}_\theta\rangle \mid \langle x, \mathbf{b}_\theta\rangle > 0,\ |\langle x, \mathbf{l}_\theta\rangle| < r_a\}\Big),
  \label{eq:freespace}
\end{equation}
where $\mathcal{O}_t$ is the set of obstacle points in the agent frame and $\mathbf{b}_\theta$, $\mathbf{l}_\theta$ are the forward and lateral unit vectors of heading $\theta$.
Because $f_t$ applies the same corridor test as \mt, the profile states which headings admit an executable move of a given length.
The \sketch adds about 350 tokens to the packet, \eg, ``\texttt{view 3 RIGHT: 1.1/4.1/1.0 \ldots}'' for the grid and ``\texttt{right 3.5\,m, 60 deg right 1.1\,m, \ldots}'' for the profile.
Polar free-space representations are standard in obstacle avoidance~\citep{borenstein1991vector}, and VLM navigators project navigable directions onto the image~\citep{goetting2024end}; the \sketch differs in its purpose, which is to anticipate the predictable read-only queries of the API VLM model so that they need not be issued.

\subsection{Intent-preserving waypoint repair}
\label{sec:repair}

The motion check regularly rejects a \mt selected by the \decider, typically because the target pixel lies just behind a door frame or too close to an obstacle.
Such rejections are geometric: the intended heading $\theta^\ast$, \ie, the bearing of the rejected pixel, is usually right, but the target point cannot be reached along a straight corridor.
Instead of paying for another API round, the harness hands the rejected call, the \decider's reason, and the rejection message to the \explorer, which proposes an executable move whose heading deviates from $\theta^\ast$ by at most $45^\circ$; every proposal passes the same collision check as \mt.
The \explorer either re-targets in the views, choosing a nearer floor point or one shifted toward open floor within up to three rounds, or selects one of the collision-checked headings that the harness enumerates on the $15^\circ$ turning grid; our implementation exposes the two as a configuration option.
If no executable move in the intended direction is found, the rejection is returned to the \decider.
The \explorer thus chooses how to move, but never where to go.

\subsection{Implementation}
\label{sec:impl}

We use Qwen3.8-27B~\cite{qwen38} as the \explorer, served locally in bf16 with vLLM~\cite{kwon2023efficient} on two GPUs and queried without extended reasoning.
Since Gemini-3.7-Flash, which was used in the baseline method, has been discontinued, we select Gemini-3.8-Flash as the primary \decider for economic reasons; in addition, we evaluate GPT-5.5, and GPT-6-Astra as \decider.
Unless otherwise specified, all \decider{}s run with the provider's default reasoning setting and temperature; 
The read budget is $K=3$.
All other harness settings (four $90^\circ$ views at $640\times480$, $\rho_{\max}=3$\,m, $m=0.5$\,m, $r_a=0.1$\,m, $15^\circ$ / $0.25$\,m discretization, at most 40 decision steps and 500 simulator steps) follow  AgenticNav \citep{li2026agenticnav}. Further implementation details are given in supplementary material.

%% file: sec/4_experiments.tex
\section{Experiments}
\label{sec:exp}

\subsection{Setup}
\label{sec:setup}

\textbf{Benchmark and metrics.}
We evaluate on R2R-CE~\citep{krantz2020beyond} in the Habitat simulator~\citep{savva2019habitat} with Matterport3D scenes~\citep{chang2017matterport3d}.
Following prior agentic work~\citep{qiao2025open,shi2025smartway,li2026agenticnav, dai2026gpt}, we use the Rand100 protocol (also called R2R-CE-100), the fixed set of 100 val-unseen episodes sampled by Open-Nav~\citep{qiao2025open}, which covers 10 scenes.
We report success rate (SR, stop within 3\,m of the goal), success weighted by path length (SPL)~\citep{anderson2018evaluation}, oracle success rate (OSR), normalized dynamic time warping (nDTW)~\citep{ilharco2019general}, and navigation error (NE, meters).
For cost, we report the API cost per episode at the list price of the API model without cache discounts, and the relative API cost with respect to the API-only agent that uses the same model.

\textbf{Baselines.}
We used the reproduced of AgenticNav~\citep{li2026agenticnav} as the API‑only baseline.
AgenticNav releases neither code nor prompts, so we reimplemented its harness from the paper and wrote the prompts
ourselves.
Gemini‑3.7‑Flash, the model of the original paper, is no longer available and is now served as Gemini‑3.8‑Flash, so the
reported numbers cannot be reproduced exactly; our reproduction reaches 77.0 SR, against 76.0 reported.
We run the API‑only agent with Gemini‑3.8‑Flash, GPT‑5.5 and GPT‑6‑Astra (at maximum reasoning effort) as the single
model for all tool rounds.
\method uses the same harness, system prompt, tool definitions, observations, and action interface; it differs only in
how the tool rounds of a step are divided between the two models.


\subsection{Main results}
\label{sec:exp_main}

\begin{table}[!htbp]
\centering
\caption{\textbf{Comparison on R2R-CE val-unseen (Rand100 / R2R-CE-100: Sampled 100 episodes by Open-Nav).} – denotes unavailable or unverified; $\pm$ denotes the standard deviation over repeated runs where reported, and evaluation settings (views, backbones) differ across methods.
Pano.: panoramic RGB(-D) views; F3+P: three frontal views plus a panorama-free design~\citep{shi2026fast}.
\textbf{Bold}: best; \underline{underline}: second best.}
\label{tab:main}
\small
\setlength{\tabcolsep}{2.8pt}
\renewcommand{\arraystretch}{1.10}
\begin{tabularx}{\linewidth}{@{}Y l l ccccc@{}}
\toprule
Method & Backbone & View & NE$\downarrow$ & OSR$\uparrow$ & SR$\uparrow$ & SPL$\uparrow$ & nDTW$\uparrow$ \\
\midrule
MapGPT-CE~\citep{chen2024mapgpt} & GPT-4o & Pano. & 8.2 & 21.0 & 7.0 & 5.0 & -- \\
DiscussNav~\citep{long2024discuss} & GPT-4 & Pano. & 7.8 & 15.0 & 11.0 & 10.5 & -- \\
Open-Nav~\citep{qiao2025open} & GPT-4 & Pano. & 6.7 & 23.0 & 19.0 & 16.1 & -- \\
Fast-SmartWay~\citep{shi2026fast} & GPT-4o & F3+P & 7.7{\scriptsize$\pm$0.42} & -- & 27.8{\scriptsize$\pm$2.22} & 25.0{\scriptsize$\pm$2.70} & -- \\
SmartWay~\citep{shi2025smartway} & GPT-4o & Pano. & 7.0 & 51.0 & 29.0 & 22.5 & -- \\
EvoNav~\citep{dai2026history} & GPT-4o & Pano. & 6.0 & 35.0 & 30.0 & 24.9 & -- \\
Three-Step Nav~\citep{zheng2026three} & GPT-5 & Pano. & 5.9 & 39.0 & 34.0 & 29.1 & -- \\
STRIDER~\citep{he2026strider} & GPT-4o & Pano. & 6.9 & 39.0 & 35.0 & 30.3 & -- \\
LaViRA~\citep{ding2026lavira} & GPT-4o & Pano. & 6.4{\scriptsize$\pm$0.28} & 43.3{\scriptsize$\pm$3.2} & 36.0{\scriptsize$\pm$1.7} & 28.3{\scriptsize$\pm$0.8} & -- \\
LaViRA~\citep{ding2026lavira} & Gemini-2.5-Pro & Pano. & 6.5{\scriptsize$\pm$0.27} & 48.7{\scriptsize$\pm$2.1} & 38.3{\scriptsize$\pm$0.6} & 28.3{\scriptsize$\pm$0.9} & -- \\
EvoNav~\citep{dai2026history} & Gemini-2.5-Pro & Pano. & 5.0 & 51.0 & 43.0 & 37.8 & -- \\
SmartWay~\citep{shi2025smartway} & GPT-5.5 & Pano. & 5.2 & 60.0 & 44.0 & 35.0 & -- \\
AgenticNav~\citep{li2026agenticnav} & Gemini-2.5-Pro & Pano. & 5.9 & 63.0 & 49.0 & 33.2 & -- \\
O2C-Nav~\citep{pan2026one} & Gemini-2.5-Pro & Pano. & 5.8 & 65.0 & 49.3 & 30.9 & -- \\
AgenticNav~\citep{li2026agenticnav} & GPT-5.5 & Pano. & 5.2 & 65.0 & 55.0 & 48.4 & -- \\
HarnessVLN~\citep{chen2026harnessvln} & GPT-5.5 & Pano. & 4.2 & 77.0 & 58.0 & 35.8 & 46.7 \\
AgenticNav~\citep{li2026agenticnav} & Gemini-3.7-Flash & Pano. & \underline{3.5} & 78.0 & 76.0 & \textbf{66.5} & 69.4 \\
AgenticNav$^\dagger$~\citep{li2026agenticnav} & Gemini-3.8-Flash & Pano. &3.8& 81.0 & 77.0 & 61.0 & \underline{69.9} \\
\midrule

\method & \makecell[l]{Gemini-3.8-Flash} & Pano. & 3.6 & \underline{86.0} & \underline{79.0} & 57.7 & 63.9 \\
\method & \makecell[l]{GPT-5.5} & Pano. & 5.1 & 59.0 & 54.0 & 45.6 & 59.0 \\
\method & \makecell[l]{GPT-6-Astra$^\ddagger$} & Pano. & \textbf{2.8} & \textbf{88.0} & \textbf{86.0} & \underline{66.1} & \textbf{70.5} \\
\bottomrule
\end{tabularx}
\vspace{2pt}
\par\footnotesize\raggedright $^\dagger$ Reproduction using the methods described in AgenticNav~\citep{li2026agenticnav}.
$^\ddagger$ At maximum reasoning effort; this is because we observed that GPT-6-Astra makes decisions directly without performing nearly any reasoning at the default.
\end{table}

\textbf{Comparison with SOTA methods.}
As shown in \cref{tab:main}, \method remains competitive with recent agentic VLN methods while assigning much of the evidence gathering to a local model.
Its Gemini-3.8-Flash configuration reaches 79.0 SR and 86.0 OSR, exceeding all baseline entries on these two metrics.
This corresponds to a 2.0-point SR increase over our reproduced API-only AgenticNav, a pattern also observed with GPT-5.5 (54.0 \vs 52.0 SR).
Neither paired SR difference is statistically significant, but these configurations reduce API cost at list prices by 45\% and 50\%, respectively.
The main trade-off for the Gemini configuration lies in path quality: its lower SPL and nDTW indicate that successful navigation does not necessarily translate into efficient routes or close adherence to the reference trajectory.
Using GPT-6-Astra at maximum reasoning effort as the \decider further raises SR to 86.0, alongside 88.0 OSR, 2.8\,m NE, and 70.5 nDTW, yielding the best results on these four metrics among the listed methods.
Its SPL of 66.1 is also close to the best reported value of 66.5, showing that the framework can attain high success together with strong path efficiency.
The last result characterizes the performance attainable with a stronger \decider.

\subsection{API cost}
\label{sec:exp_cost}

\begin{table}[!htbp]
\centering
\caption{\textbf{API cost with different \decider{}s.} API cost per episode at the list price of each model without cache discounts (per million input / output tokens: Gemini-3.8-Flash \$0.75 / \$3.75, GPT-5.5 \$5 / \$30, GPT-6-Astra \$10 / \$50); Rel.: API cost relative to the API-only agent with the same model.}
\label{tab:cost_deciders}
\setlength{\tabcolsep}{3.5pt}
\small
\begin{tabular*}{\linewidth}{@{\extracolsep{\fill}} l l c c c c @{} }
\toprule
\decider model & Agent & SR$\uparrow$ & API calls$\downarrow$ & Cost$\downarrow$ & Rel.$\downarrow$ \\
\midrule
\multirow{2}{*}{Gemini-3.8-Flash} & API-only & 77.0 & 53.5 & 0.600 & 100\% \\
 & \method & \textbf{79.0} & \textbf{33.2} & \textbf{0.328} & \textbf{54.7\%} \\
\midrule
\multirow{2}{*}{GPT-5.5} & API-only & 52.0 & 16.7 & 0.565 & 100\% \\
 & \method & \textbf{54.0} & \textbf{7.6} & \textbf{0.282} & \textbf{49.9\%} \\
\midrule
\multirow{2}{*}{GPT-6-Astra} & API-only & 82.0 & 20.7 & 1.873 & 100\% \\
 & \method & \textbf{86.0} & \textbf{11.0} & \textbf{0.871} & \textbf{46.5\%} \\
\bottomrule
\end{tabular*}
\end{table}

\textbf{Savings across \decider{}s.}
\Cref{tab:cost_deciders} compares the API cost per episode of \method with that of the API-only agent using the same model.
With Gemini-3.8-Flash, GPT-5.5 and GPT‑6‑Astra as \decider, \method reduces the API cost (based on the officially published list price) by nearly half.
Although the input costs of these three models vary significantly, their relative efficiency is roughly comparable.
OpenAI consumes far fewer API tokens than Gemini because it counts a single view as approximately 300 tokens, resulting in a context length for the same number of steps that is less than half; furthermore, we found that it collects fewer pieces of evidence at each decision step—particularly by rarely looking back at past views—which further reduces the number of evidence packets. Another reason is that OpenAI’s automatic prompt caching mechanism further reduces the lengthy, repetitive prefixes found in pure API proxies.

\begin{table}[!htbp]
\centering
\caption{\textbf{Where the API cost goes.} Share of the API cost and API calls per episode by the tool that ends a round: information (\qd, \rc) or action (\mt, \st).
In the API-only agent, information rounds dominate; in \method most of them run on the local \explorer, and the remaining API cost is dominated by the \decider's recalls of past views.}
\label{tab:cost}
\small
\setlength{\tabcolsep}{4.0pt}
\begin{tabular}{l cc cc}
\toprule
 & \multicolumn{2}{c}{API-only} & \multicolumn{2}{c}{\method} \\
\cmidrule(lr){2-3}\cmidrule(lr){4-5}
Round type & share & calls/ep & share & calls/ep \\
\midrule
Information & 78.3\% & 49.4 & 62.3\% & 22.1 \\
\quad of which \qd & 37.3\% & 22.2 & 21.6\% & 6.9 \\
\quad of which \rc & 41.0\% & 27.2 & 40.7\% & 15.2 \\
Action & 21.7\% & 10.9 & 37.7\% & 11.1 \\
\midrule
API \$/ep & \multicolumn{2}{c}{0.600} & \multicolumn{2}{c}{0.328} \\
\bottomrule
\end{tabular}
\end{table}

\textbf{What drains the API budget?}
\Cref{tab:cost} breaks the API cost down by the tool that ends each round.
In the API-only agent, 78.3\% of the cost is spent on rounds that only gather information, split almost evenly between recalls of past views (41.0\%) and depth queries (37.3\%): each such round re-sends the four views, the BEV map, and the step's previous rounds, so its input cost is as high as that of an action round.
\method moves most of these rounds to the local \explorer, which issues 27.1 calls per episode.
Depth rounds of the API model drop from 22.2 to 6.9 per episode, because the \explorer's readings and the \sketch answer most depth questions in advance; recall rounds drop from 27.2 to 15.2.
The Decider’s post-handoff recall calls still account for 40.7\% of its API cost.

\subsection{Ablation study}
\label{sec:ablation}

\begin{table}[!htbp]
\centering
\caption{\textbf{Component ablation.}
Top: Qwen3.8‑27B as both \explorer and \decider, which isolates the components at zero API cost.
Bottom: the same ablations with Gemini‑3.8‑Flash as \decider.
Call: \decider calls per episode, a proxy for the API load.}
\label{tab:ablation}

\small
\renewcommand{\arraystretch}{1}
\setlength{\tabcolsep}{3.5pt}

\begin{tabular*}{\linewidth}{@{\extracolsep{\fill}}lcccc@{}}
\toprule
Variant & SR$\uparrow$ & SPL$\uparrow$ & nDTW$\uparrow$ & Call$\downarrow$ \\
\midrule
\multicolumn{5}{@{}l@{}}{\emph{\decider{}: Qwen3.8-27B}} \\
\method{} (full)          & 58.0 & 40.7 & 53.5 & 16.2 \\
\quad w/o \sketch         & 53.0 & 37.4 & 52.1 & 20.7 \\
\quad w/o waypoint repair & 60.0 & 42.3 & 55.3 & 16.4 \\
\quad w/o \packet         & 57.0 & 38.4 & 49.8 & 27.3 \\
\midrule
\multicolumn{5}{@{}l@{}}{\emph{\decider{}: Gemini-3.8-Flash}} \\
\method{} (full)          & 79.0 & 57.7 & 63.9 & 33.2 \\
\quad w/o \sketch         & 77.0 & 55.9 & 62.8 & 35.9 \\
\quad w/o waypoint repair & 69.0 & 51.3 & 60.5 & 33.5 \\
\quad w/o \packet         & 73.0 & 47.7 & 51.4 & 44.1 \\
\bottomrule
\end{tabular*}
\end{table}

We remove (i) the \sketch, (ii) the waypoint repair, and (iii) the \packet itself, including the \sketch, in which case
the \decider receives the \explorer's full tool conversation of the step.
To isolate the contribution of each component without API cost, we first run the ablations with Qwen3.8‑27B as both
\explorer and \decider; the \decider uses extended reasoning (thinking mode at its default effort).
We then repeat them with Gemini‑3.8‑Flash, the results are shown in \cref{tab:ablation}.

\textbf{Look--act decoupling.}
Even when both roles use the same model, decoupling halves the load of the reasoning model at equal success: the \decider of \method issues 16.2 calls with 132k input tokens per episode, against 27.6 calls and 266k tokens for the single model (58.0 \vs 57.0 SR).
The saving not depend on the cost gap between a local and an API model; it comes from moving read-only rounds to a model that is queried without extended reasoning.

\textbf{Evidence packet.}
Replacing the \packet with the \explorer's full tool conversation raises the \decider's load with both \decider{}s, from 16.2 to 27.3 calls per episode with Qwen3.8-27B and from 33.2 to 44.1 with Gemini-3.8-Flash, without improving success (57.0 \vs 58.0 and 73.0 \vs 79.0 SR).
It also lowers path quality: SPL drops from 40.7 to 38.4 and from 57.7 to 47.7, and nDTW from 53.5 to 49.8 and from 63.9 to 51.4.
The compact packet thus transfers the \explorer's evidence at a lower cost than its conversation and leads to more efficient paths.

\textbf{Scene sketch.}
Removing the \sketch raises the \decider's load from 16.2 to 20.7 calls per episode with Qwen3.8-27B and from 33.2 to 35.9 with Gemini-3.8-Flash, and lowers all three navigation metrics with both \decider{}s: SR from 58.0 to 53.0 and from 79.0 to 77.0, SPL from 40.7 to 37.4 and from 57.7 to 55.9, and nDTW from 53.5 to 52.1 and from 63.9 to 62.8.
This matches the motivation of \cref{sec:packet}: the \sketch answers in advance the depth questions that the \decider would otherwise pay for, saving \decider calls while improving navigation.

\textbf{Waypoint repair.}
With the open-weight \decider, removing the repair leaves the load unchanged (16.4 \vs 16.2 calls per episode) and changes the navigation metrics by at most two points (60.0 \vs 58.0 SR, 42.3 \vs 40.7 SPL, 55.3 \vs 53.5 nDTW).
With Gemini-3.8-Flash, removing it lowers SR from 79.0 to 69.0, SPL from 57.7 to 51.3, and nDTW from 63.9 to 60.5 at nearly the same load (33.5 \vs 33.2 calls per episode).
The repair therefore matters most with the stronger \decider, whose waypoints are worth preserving once their direction has been chosen.

\subsection{Real-world Experiments}
\label{sec:real_world}

\textbf{Deployment setup.}
We deploy \method on Unitree Go2 EDU quadruped and Unitree G1 humanoid robots without task-specific training or fine-tuning.
A RealSense D455 RGB-D camera provides metric depth, supplemented by onboard LiDAR, while an Insta360 panoramic camera provides four $90^\circ$ perspective views at each decision step.
An onboard NVIDIA Jetson Orin handles sensing, speech recognition and motion execution.
The harness and Qwen3.8-27B \explorer run on a local NVIDIA A800 (80\,GB) server for evidence gathering and waypoint repair; only Gemini-3.8-Flash, the \decider, is accessed through a cloud API to select actions.

\begin{figure}[!htbp]
  \centering
  \includegraphics[width=\linewidth,keepaspectratio]{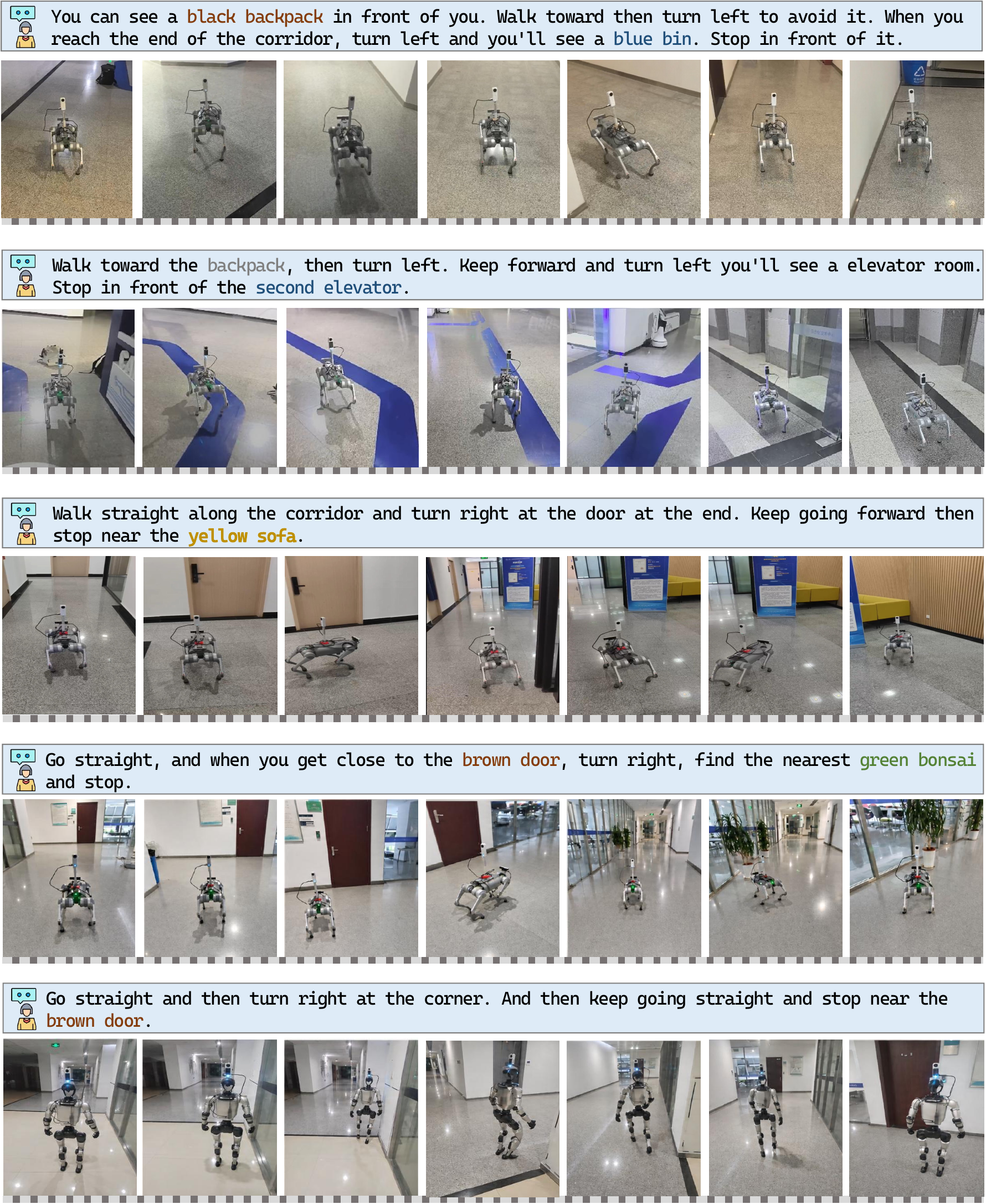}
  \caption{\textbf{Zero-shot real-world navigation with \method.} Four Unitree Go2 EDU rollouts (top) and one Unitree G1 rollout (bottom), each showing the instruction and sequential execution frames. It
can accurately execute complex natural language commands involving spatial reasoning and object
localization, demonstrating robust performance and excellent generalization from simulation to
real-world scenarios even under challenging conditions.}
  \label{fig:real}
\end{figure}

\textbf{Qualitative results.}
As shown in \cref{fig:real}, both robots follow spoken instructions in office environments with diverse layouts and illumination.
Four Go2 rollouts cover obstacle avoidance and landmark-based stopping at a blue bin, the second elevator, a yellow sofa and a green plant; the G1 follows a corridor and stops near a brown door.
These examples illustrate zero-shot transfer across robot embodiments and physical environments, including landmark grounding, spatial reasoning and instruction-conditioned stopping.

\subsection{How do \decider{}s use the \packet?}
\label{sec:analysis}

\begin{figure}[!htbp]
  \centering
  \includegraphics[width=\linewidth]{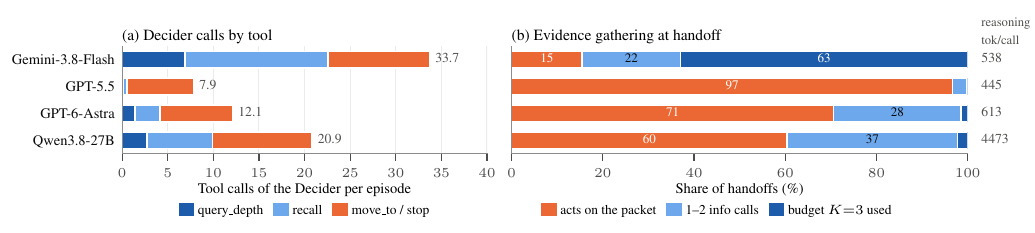}
  \caption{\textbf{Behavior of different \decider{}s in \method.}
  (a) Tool calls of the \decider per episode; a round may contain several calls.
  (b) Share of handoffs in which the \decider acts on the \packet directly, after one or two information calls, or only after using its budget of $K=3$ information rounds; right: reasoning tokens per \decider call.}
  \label{fig:decider}
\end{figure}

\textbf{Where the API calls go.}
\Cref{fig:decider}(a) shows which tools the \decider calls.
Compared with the API-only agent of the same model, \method leaves the action calls almost unchanged (11.1 \vs 10.9 \mt and \st calls per episode with Gemini-3.8-Flash, 7.3 \vs 9.0 with GPT-5.5) and removes most information calls from the API model (22.6 \vs 50.8 with Gemini-3.8-Flash, 0.6 \vs 8.4 with GPT-5.5).
Information gathering thus moves to the \explorer, while the \decider keeps the state-changing actions. This decoupled architecture has the potential to significantly reduce agent API costs.

\textbf{\decider{}s differ in evidence seeking.}
\Cref{fig:decider}(b) shows how much additional evidence each \decider requests at handoff.
GPT-5.5 acts on the \packet directly in 97\% of the handoffs and GPT-6-Astra in 71\%, requesting one or two depth values or past views otherwise.
Gemini-3.8-Flash requests additional evidence in 85\% of the handoffs, mostly recalls of past views (69\% of its information calls), and uses its full budget in 63\%.
The API \decider{}s generate comparable numbers of reasoning tokens, suggesting that a model’s preferences are related to its training architecture, and that higher-performance models do not necessarily perform better on the VLN task.

\subsection{What compromises path efficiency?}
\label{sec:supp_path}

\begin{table}[!htbp]
\centering
\caption{\textbf{Path efficiency with Gemini-3.8-Flash as \decider.}
Path: path length per episode (m); Forced: share of the \decider's moves issued after its information budget is spent (\%); Rev.: moves that reverse the heading by $\geq 135^\circ$ (\%); Osc.: pairs of consecutive reversals per episode; \$/ep: API cost per episode.
Visual history adds the start panorama and the previous front view to the \packet ($K=4$); reversal confirmation asks the \decider to confirm a second consecutive reversal.}
\label{tab:path}
\small
\setlength{\tabcolsep}{2pt}
\begin{tabular*}{\columnwidth}{@{\extracolsep{\fill}} l ccccccc}
\toprule
 & SR & SPL & Path & Forced & Rev. & Osc. & \$/ep \\
\midrule
API-only & 77.0 & 61.0 & 15.8 & -- & 19.2 & 0.37 & 0.600 \\
\method & 79.0 & 57.7 & 19.8 & 70.9 & 24.6 & 0.93 & 0.328 \\
\quad + visual history & 76.0 & 60.9 & 16.6 & 42.4 & 20.2 & 0.36 & 0.401 \\
\quad + reversal confirm & 74.0 & 55.1 & 18.8 & 70.0 & 23.3 & 0.47 & 0.310 \\
\bottomrule
\end{tabular*}
\end{table}

With Gemini-3.8-Flash as \decider, \method succeeds as often as the API-only agent but travels longer paths, reverses its heading more often, and oscillates more, which lowers SPL from 61.0 to 57.7.
\Cref{tab:path} shows that the reason is a mandatory detour: the \decider makes 70.9\% of its moves only after it has spent its $K=3$ information calls, mostly on recalls of past views that the \packet does not contain, and these forced moves reverse the heading in 31.0\% of the cases, against 12.3\% for its voluntary moves.
Supplying this visual history in the \packet (start panorama and previous front view, $K=4$) reduces the forced moves to 42.4\%, brings reversals and oscillations to the level of the API-only agent, and restores SPL to 60.9 at \$0.401 per episode, whereas asking the \decider to confirm a second consecutive reversal leaves the forced moves at 70.0\% and lowers SPL to 55.1; the \decider thus lacks visual context rather than deliberation.

%% file: sec/5_conclusion.tex
\section{Conclusion}
\label{sec:conclusion}

We presented \method, an agentic framework for cost-efficient vision-and-language navigation in continuous environments.
Its central idea, look--act decoupling, allows a local model to gather evidence for a stronger API model's navigation decisions.
Experiments show that this method achieves competitive success rates across a variety of backbone networks, with an API cost nearly half that of corresponding pure API baseline models.
This method promotes the joint design of model harness collaboration and visual memory to build more efficient embodied agents. 

%% file: sec/X_suppl.tex
\clearpage
\section*{Appendix}

\section{Implementation details}
\label{sec:supp_impl}

\textbf{Harness.}
We reimplemented its harness from the AgenticNav~\citep{li2026agenticnav}.
\Cref{tab:supp_harness} lists the settings that the paper leaves open and the values we use for all agents.

\begin{table}[!htbp]
\centering
\caption{Harness settings shared by all agents.}
\label{tab:supp_harness}
\small
\begin{tabular}{@{}l p{0.70\columnwidth}@{}}
\toprule
Views & 4 $\times$ $90^\circ$ HFOV (front, left, back, right), $640\times480$, camera 1.25\,m \\
Motion & $\rho_{\max}=3$\,m, stop margin $m=0.5$\,m, corridor half-width $r_a=0.10$\,m \\
Discretization & $15^\circ$ turns, $0.25$\,m forward steps, abort after 2 stalled steps \\
Budgets & 40 decision steps, 500 simulator steps, 10 tool rounds per step \\
BEV map & 0.1\,m grid, ego-centric $16\times16$\,m crop at 512\,px \\
History & reasons and actions of the last 6 steps \\
\bottomrule
\end{tabular}
\end{table}

\textbf{Models.}
\explorer: Qwen3.8-27B in bf16, served with vLLM 0.17.1 with tensor parallelism over NVIDIA A800 80GB GPU and queried without extended reasoning.
\decider{}s: Gemini-3.8-Flash through the Gemini API, GPT-5.5 through the OpenAI Chat Completions API, and GPT-6-Astra through the OpenAI Responses API, the only OpenAI interface that combines function tools with reasoning for this model.
Unless otherwise specified, all \decider{}s use the provider's default inference settings and temperature parameters. Within a decision step, a \decider receives its own earlier reasoning where the interface returns it; the Chat Completions interface of GPT-5.5 does not return it.
Ablations with an open-weight \decider use Qwen3.8-27B in thinking mode at its default effort.

\section{Capacity of the \explorer}
\label{sec:supp_scout}

\begin{table}[!htbp]
\centering
\caption{\textbf{Capacity of the \explorer.} \decider: Gemini-3.8-Flash. Calls: \decider calls per episode; \$/ep: API cost per episode; Rel.: API cost relative to the API-only agent; $p$: paired McNemar test on SR against the Qwen3.8-27B \explorer.}
\label{tab:supp_scout}
\small
\setlength{\tabcolsep}{2.5pt}
\begin{tabular*}{\columnwidth}{@{\extracolsep{\fill}} l ccc cccc}
\toprule
\explorer & SR & SPL & nDTW & Calls & \$/ep & Rel. & $p$ \\
\midrule
Qwen3.8-27B & 79.0 & 57.7 & 63.9 & 33.2 & 0.328 & 54.7\% & -- \\
Qwen3.5-9B & 81.0 & 60.4 & 65.6 & 27.6 & 0.275 & 45.8\% & 0.82 \\
Qwen3.5-4B & 75.0 & 57.6 & 63.7 & 31.2 & 0.292 & 48.7\% & 0.50 \\
\midrule
API-only & 77.0 & 61.0 & 69.9 & 53.5 & 0.600 & 100\% & -- \\
\bottomrule
\end{tabular*}
\end{table}

\begin{table}[!htbp]
\centering
\caption{\textbf{Behavior with different \explorer{}s.} Rounds: \explorer rounds per decision step; Depth: depth points measured by the \explorer per handoff; Info: information calls of the \decider per handoff; Forced: share of the \decider's moves issued after its budget is spent (\%); Rev.: moves that reverse the heading by $\geq 135^\circ$ (\%); Path: path length per episode (m); Stop: handoffs triggered by an intended \st (\%).}
\label{tab:supp_scout_behavior}
\small
\setlength{\tabcolsep}{2.5pt}
\begin{tabular*}{\columnwidth}{@{\extracolsep{\fill}} l ccc ccc c}
\toprule
\explorer & Rounds & Depth & Info & Forced & Rev. & Path & Stop \\
\midrule
Qwen3.8-27B & 2.12 & 2.9 & 2.21 & 70.9 & 24.6 & 19.8 & 9.7 \\
Qwen3.5-9B & 3.04 & 6.6 & 2.03 & 63.6 & 20.3 & 17.3 & 9.1 \\
Qwen3.5-4B & 2.12 & 4.0 & 2.21 & 70.8 & 21.8 & 18.3 & 5.1 \\
\bottomrule
\end{tabular*}
\end{table}

\textbf{Setup.}
Qwen3.8 has no smaller open-weight model, so we replace the Qwen3.8-27B \explorer by Qwen3.5-9B and Qwen3.5-4B, which share its architecture but belong to the previous generation.
Each is served in bf16 by vLLM on a single A800 GPU, with 19\,GB and 9\,GB of weights.
All other settings are those of the main experiment: Gemini-3.8-Flash as \decider, the \packet with the \sketch, $K=3$, waypoint repair by re-targeting, and the same 100 episodes, with one run per \explorer.

\textbf{Results.}
With both smaller \explorer{}s, the API cost stays at about half of that of the API-only agent (45.8\% and 48.7\%, \cref{tab:supp_scout}).
Qwen3.5-9B matches the success of Qwen3.8-27B (81.0 \vs 79.0 SR, +11 / $-$9 episodes, $p=0.82$), and Qwen3.5-4B reaches 75.0 SR (+8 / $-$12 episodes, $p=0.50$).
The API cost is set by the \decider: every decision step ends in a handoff, and the \decider's input is dominated by the observations rendered by the harness, so the \decider issues 3.2--3.4 calls with about 10k input tokens per handoff with all three \explorer{}s.

\textbf{Behavior Analysis.}
Qwen3.5-9B gathers more evidence before handing off than Qwen3.8-27B: it runs 3.04 instead of 2.12 rounds per step and measures 6.6 instead of 2.9 depth points per handoff.
With this evidence, the \decider requests fewer additional observations and makes fewer forced moves (63.6\% \vs 70.9\%); the agent reverses its heading less often and travels shorter paths (17.3 \vs 19.8\,m), which raises SPL to 60.4 and lowers the API cost to \$0.275 per episode.
This is consistent with the findings of path analysis: the more evidence the \decider receives at handoff, the fewer moves it makes in the forced round.
Qwen3.5-4B hands off with an intended \st about half as often as the larger \explorer{}s (5.1\% of its handoffs). The experimental results above indicate that the model’s capacity gains are not significant and are more closely related to the training framework and method used; therefore, future work could involve training an open-source Scout model as an alternative to the Qwen.

\section{Where does the saving come from?}
\label{sec:supp_saving}

Compared with the API-only agent, \method changes two things: it adds the local \explorer, and it consults the API model only to select an action, with a bounded read budget and a compact \packet.
To separate the two, we remove the \explorer's evidence loop and keep everything else (\emph{w/o \explorer evidence}).
Every step is handed to the \decider at round 0 with the packet that the harness builds without any model call, \ie, the views, the BEV map, the history, and the \sketch, whose floor depth grid already measures 36 fixed floor pixels and thus serves as a rule-based prefetch of depth.
The \decider (Gemini-3.8-Flash), the read budget $K=3$ with a forced round that offers only \mt, and waypoint repair by the \explorer are unchanged, and the variant runs once on the same 100 episodes.

\begin{table}[!htbp]
\centering
\caption{\textbf{Decoupled protocol}.
\emph{w/o \explorer evidence} keeps the protocol (handoff for actions only, $K=3$, \packet with \sketch, repair) but hands every step to the \decider without any \explorer reads.
Calls, input tokens, and cost per episode refer to the API model.}
\label{tab:supp_saving}
\small
\setlength{\tabcolsep}{2pt}
\begin{tabular*}{\columnwidth}{@{\extracolsep{\fill}} l ccc ccc}
\toprule
 & SR & SPL &  Calls & Input & \$/ep  \\
\midrule
API-only & 77.0 & 61.0 &  53.5 & 664k & 0.600  \\
\midrule
w/o \explorer evidence & 72.0 & 51.8 &  36.8 & 357k & 0.329  \\
w/o \explorer (only geometry) & 73.0 & 55.4 &  34.4 & 343k & 0.331  \\
\method & 79.0 & 57.7 &  33.2 & 341k & 0.328  \\
\bottomrule
\end{tabular*}
\end{table}

\textbf{The protocol provides the saving.}
Without the \explorer's evidence, the API cost per episode equals that of \method, and the API model issues 36.8 instead of 53.5 calls per episode (\cref{tab:supp_saving}).
The saving therefore comes from the decoupled protocol: the API model is consulted only to select an action, may add at most $K$ reads, and starts each step from a compact \packet in which the \sketch answers the most frequent depth questions in advance.
Its depth rounds drop from 22.2 per episode in the API-only agent to 9.1 without the \explorer's readings and to 6.9 with them.
The saving also holds across local models: with Qwen3.5-9B or Qwen3.5-4B as \explorer, the API cost stays between 46\% and 55\% of that of the API-only agent in
\cref{tab:supp_scout}.

\textbf{Analysis of Scout ablation.}
The \explorer's evidence determines how well the bounded budget is spent.
Without it, the \decider spends more of its reads on depth (9.1 \vs 6.9 depth queries per episode), SR drops from 79.0 to 72.0, below that of the API-only agent 77.0, and SPL drops from 57.7 to 51.8.
It points in the same direction as the repair ablation, in which removing the repair performed by the \explorer lowers SR from 79.0 to 69.0. In addition, we also attempt to completely remove Scout and replace its induction with geometric techniques commonly used in navigation algorithms; the results show that the SR still decreases.
Together, these results suggest a division of labor: look--act decoupling provides the saving, and the \explorer's evidence and repair keep the success of the bounded \decider at the level of the API-only agent.
\vspace{-0.5cm}
\section{Bounded commitment}
\label{sec:supp_forced}

When the \decider has spent its $K$ reads, it has usually stopped extracting evidence from the current position: 78\% of the reads that precede a forced commitment recall past views rather than measure the current ones.
\method then forces a move toward the \decider's most promising waypoint, which makes progress and yields a new viewpoint with fresh evidence at the next step.
The forced round offers only \mt, because a move can be corrected at the next step, whereas \st ends the episode.
\Cref{tab:supp_forced} compares this design with a preliminary harness that also offered \st in the forced round.

\begin{table}[!htbp]
\centering
\caption{\textbf{Forced round with and without \st} (Gemini-3.8-Flash \decider, Rand100). Forced stops: episodes that end with a \st issued in the forced round. The preliminary harness did not yet include the \sketch.}
\label{tab:supp_forced}
\small
\setlength{\tabcolsep}{3pt}
\begin{tabular*}{\columnwidth}{@{\extracolsep{\fill}} lccc}
\toprule
Harness & Forced round & Forced stops & SR \\
\midrule
Preliminary & \mt, \st & 63 & 46.0 \\
Final & \mt & 0 & 79.0 \\
\midrule
API-only agent & -- & -- & 77.0 \\
\bottomrule
\end{tabular*}
\end{table}

When \st is offered in the forced round, the \decider ends 63 of the 100 episodes with it, at a median of decision step 2 and 5.6\,m from the goal, and SR falls to 46.0; these episodes include 32 of the 34 that the API-only agent solves and the preliminary run fails.
The reasons attached to these calls show no intention to stop, \eg, \emph{``I am not done yet, wait don't stop!''}: an API model that must act before it is ready tends to take the terminal option when it is offered.
Restricting the forced round to \mt removes these premature stops, and the final configuration reaches 79.0 SR, above the API-only agent 77.0.

\raggedbottom  
\providecolor{cvprblue}{rgb}{0.21,0.49,0.74}  
\newcommand{\promptbox}[2]{\par\medskip\noindent{\setlength{\fboxsep}{0pt}\setlength{\fboxrule}{0.4pt}%
  \fcolorbox{cvprblue!55}{white}{\parbox{\dimexpr\linewidth-0.8pt\relax}{%
    \colorbox{cvprblue!12}{\parbox{\dimexpr\linewidth\relax}{\rule{0pt}{12pt}\hspace{8pt}\textbf{#1}\rule[-5pt]{0pt}{5pt}}}\par
    \vspace{5pt}\hspace*{8pt}\parbox{\dimexpr\linewidth-16pt\relax}{\small\setlength{\parskip}{4pt}#2}\par\vspace{8pt}}}}\par\medskip}
\definecolor{pjkeyc}{RGB}{0,84,147}
\definecolor{pjstrc}{RGB}{24,121,78}
\definecolor{pjnumc}{RGB}{190,90,10}
\definecolor{pjphc}{RGB}{110,110,110}
\newcommand{\pjkey}[1]{\textcolor{pjkeyc}{"#1"}}
\newcommand{\pjstr}[1]{\textcolor{pjstrc}{"#1"}}
\newcommand{\pjnum}[1]{\textcolor{pjnumc}{#1}}
\newcommand{\pjph}[1]{\textcolor{pjphc}{"<#1>"}}
\newcommand{\promptjson}[1]{\par\vspace{3pt}{\setlength{\fboxsep}{5pt}\setlength{\fboxrule}{0.4pt}\fcolorbox{cvprblue!45}{cvprblue!6}{\parbox{\dimexpr\linewidth-10.8pt\relax}{\ttfamily\footnotesize\frenchspacing #1}}}}
\vspace{-0.5cm}
\section{Model prompts}
\label{sec:supp_prompts}

The \explorer and the \decider share one system prompt and one tool set, written for our reimplementation of the AgenticNav harness~\citep{li2026agenticnav}; the two roles differ in when they are called and in what they receive.
Every call contains the instruction, the recent decision history, the four labeled views, the BEV map, and the tool feedback of the current step; the \decider receives these inputs through the \packet. The following excerpts are condensed for presentation; the JSON examples show selected fields.

\subsection{Prompt excerpts}

\promptbox{Shared system prompt: task, observations, and tools}{%
\textbf{Task.} Follow the navigation instruction in an unseen indoor building and stop at the described final location; an episode succeeds if \st is called within 3\,m of the goal.

\textbf{Observations.} [instruction], [last six reasons and executed actions], [views 0--3: front, left, back, right], [BEV map with the path and numbered past decision points], [step and distance counters]. Pixel coordinates are normalized, and depth is obtained only through \qd.

\textbf{Tools.} \qd returns metric depth with neighborhood statistics at up to 12 pixels per call; \rc returns view $k$ observed at past decision point $p$; \mt turns toward the back-projected point under a pixel and walks at most 3\,m, or returns \texttt{reselect} with the free distance; \st ends the episode.

\textbf{Action selection.} Work out which parts of the instruction are done and which landmark comes next, and write this as the reason, which becomes the memory of later steps. Prefer floor targets 1.5--3\,m away in open space; to pass a doorway, target the floor just beyond its frame. Stop only when every part of the instruction is complete.
\promptjson{\{\\
\hspace*{1.2em}\pjkey{name}: \pjstr{move\_to},\\
\hspace*{1.2em}\pjkey{parameters}: \{\pjkey{k}: \pjph{view index: 0 front, 1 left, 2 back, 3 right},\\
\hspace*{7.2em}\pjkey{u}: \pjph{normalized column}, \pjkey{v}: \pjph{normalized row},\\
\hspace*{7.2em}\pjkey{reason}: \pjph{completed part of the instruction and next landmark}\}\\
\}}}

\promptbox{\explorer: gather evidence before acting}{%
\textbf{Inputs.} [step observations], [tool feedback of the step].

\textbf{Evidence gathering.} Measure what the decision depends on, such as the distance to a door or the free floor in a direction, and recall past views when the route must be re-checked. Batch all depth queries of a round into one call. The \explorer is queried without extended reasoning.

\textbf{Handoff.} The first \mt or \st that the \explorer proposes is not executed; it ends the evidence loop and hands the step to the \decider.
\promptjson{\{\\
\hspace*{1.2em}\pjkey{name}: \pjstr{query\_depth},\\
\hspace*{1.2em}\pjkey{args}: \{\pjkey{points}: [\{\pjkey{k}: \pjnum{0}, \pjkey{u}: \pjnum{0.50}, \pjkey{v}: \pjnum{0.80}\}, \{\pjkey{k}: \pjnum{1}, \pjkey{u}: \pjnum{0.35}, \pjkey{v}: \pjnum{0.75}\}]\}\\
\}}}

\promptbox{\decider: decide from the evidence packet}{%
\textbf{Inputs.} [instruction], [history], [views 0--3], [BEV map], [\explorer's depth readings], [recalled views], [scene sketch].

\textbf{Evidence.} The \explorer's measurements are presented as exact data, followed by the floor depth grid of each view and the free walking distance every $30^\circ$. The \explorer's own proposal is not shown.

\textbf{Decision.} Finish the step with \mt or \st, and request further evidence only if it is essential. After $K$ reads, only \mt is offered.
\promptjson{\{\\
\hspace*{1.2em}\pjkey{name}: \pjstr{move\_to},\\
\hspace*{1.2em}\pjkey{args}: \{\pjkey{k}: \pjnum{0}, \pjkey{u}: \pjnum{0.56}, \pjkey{v}: \pjnum{0.78},\\
\hspace*{4.8em}\pjkey{reason}: \pjph{completed part of the instruction and next landmark}\}\\
\}}}

\promptbox{Waypoint repair: keep the intended direction}{%
\textbf{Inputs.} [step observations], [\decider's target and reason], [rejection with the free distance].

\textbf{Repair.} Choose an executable floor target that serves the same intent within $45^\circ$ of the \decider's direction, such as a closer floor point or one shifted toward open floor, and measure depth if needed, within three rounds. When the harness enumerates collision-checked headings instead, choose one of at most three candidates marked in the views.
\promptjson{\{\\
\hspace*{1.2em}\pjkey{name}: \pjstr{choose\_candidate},\\
\hspace*{1.2em}\pjkey{args}: \{\pjkey{index}: \pjnum{2}, \pjkey{reason}: \pjph{passes through the intended opening}\}\\
\}}}

\subsection{Execution contract}
The models only propose tool calls; the harness validates and executes them and returns the result as tool feedback.
Pixel coordinates are normalized to $[0,1]$, with $u$ from left to right and $v$ from top to bottom, and view indices refer to the current heading.\qd and \rc are read-only: they return depth statistics at the requested pixels and the views stored at past decision points, without changing the state of the agent.A \mt target is back-projected with depth, and the move is accepted only if the corridor of half-width $r_a$ along its bearing is free of obstacle points; it is then executed as $15^\circ$ turns and 0.25\,m steps, whereas a blocked target returns \texttt{reselect} with the free distance along its bearing.
A \mt or \st proposed by the \explorer is not executed but hands the step to the \decider, each tool result reports the number of remaining rounds, and after $K$ reads the \decider is offered only \mt.
\section{Simulation visualizations}
\label{sec:supp_vis}

Following HarnessVLN~\cite{chen2026harnessvln}, we illustrate six recorded rollouts of \method with Gemini-3.8-Flash as \decider from the main experiment on Rand100.
The episodes were selected to show how the components of \method act during execution, not to estimate aggregate performance.
Each of the three successful episodes highlights one mechanism.

As shown in \cref{fig:case265}, waypoint repair turns a rejected waypoint into an executable move in the same direction; \cref{fig:case387} shows the \decider overriding an early stop proposed by the \explorer; and \cref{fig:case559} shows the agent tracking a sequence of landmarks.
The three failures illustrate typical failure modes: the agent oscillates after forced moves in \cref{fig:case94}, stops at the wrong landmark in \cref{fig:case469}, and enters a neighboring room just outside the success radius in \cref{fig:case715}.
Each figure shows the recorded view of a key decision step with the \decider's waypoint (red) and, where applicable, the waypoint executed after repair by the \explorer (green), the recorded front view at the final step, and the trajectory on a top-down render of the scene (orange: \method; dashed blue: reference path; S: start; G: goal; E: episode end).

Instructions are quoted verbatim from R2R-CE, and episode numbers refer to its val-unseen split.
The trajectories are obtained by replaying the executed actions in the simulator without any model call, and the replayed views match the views saved during the run, so each render shows the executed path exactly.


\clearpage
\subsection{Successful navigation}
\begingroup
\setlength{\intextsep}{5pt}
\captionsetup[figure]{font=small,skip=4pt}

\begin{figure}[H]
\centering
\setlength{\tabcolsep}{2pt}
\begin{adjustbox}{max width=\linewidth}\begin{tabular}{ccc}
\small RGB: step 1, front view & \small RGB: step 3, stop & \small Map \\
\includegraphics[height=4.4cm]{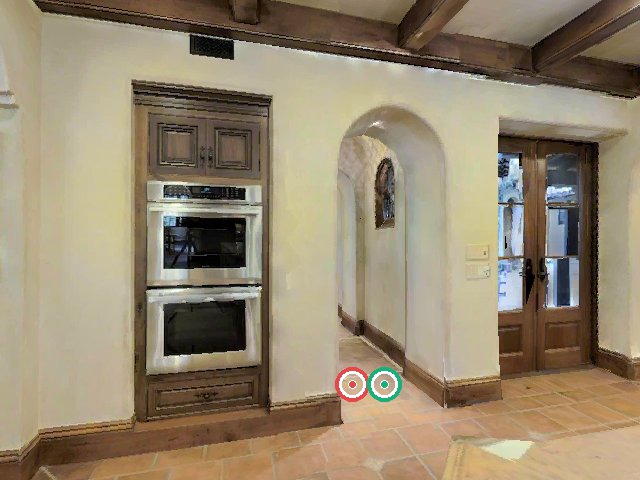} & \includegraphics[height=4.4cm]{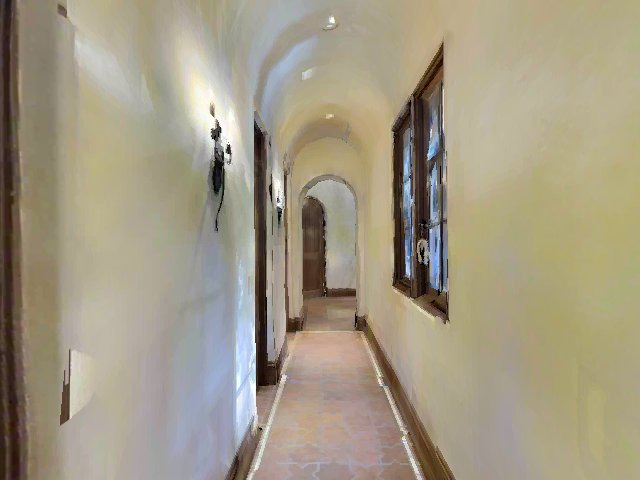} & \includegraphics[height=4.4cm]{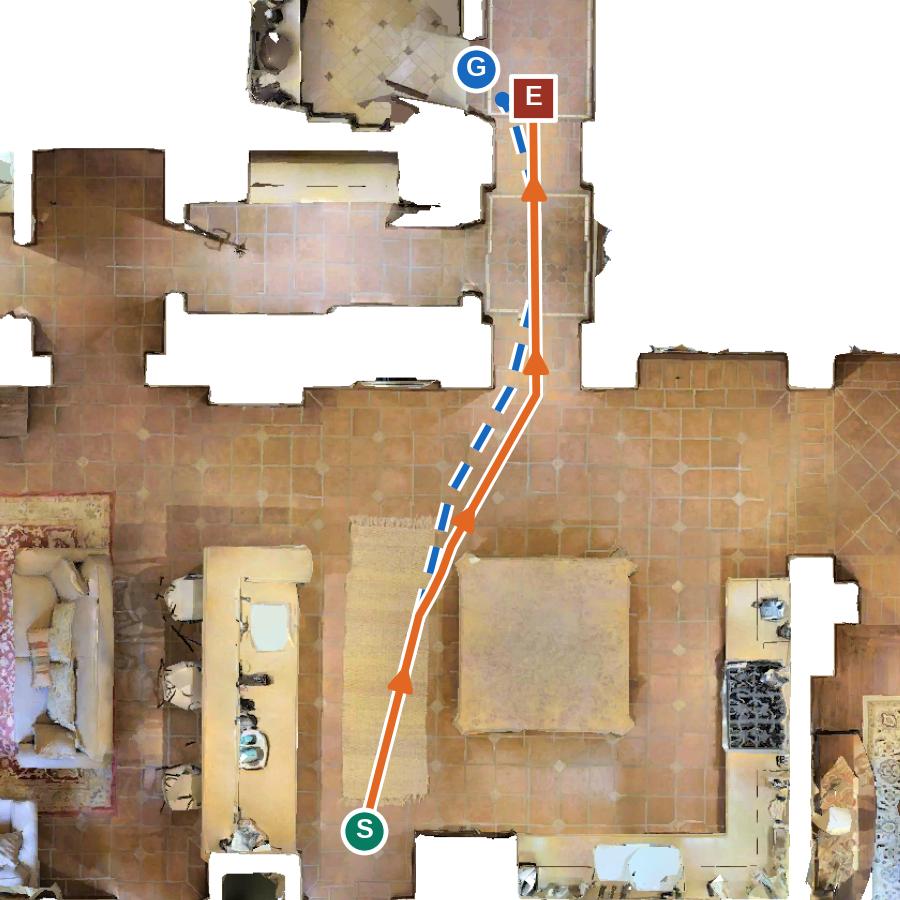} \\
\end{tabular}\end{adjustbox}
\caption{\textbf{Episode 265: waypoint repair keeps the intended route.} \emph{Walk towards oven and turn right through doorway. Go down hallway and turn left and wait at bathroom.} At step 1 the motion check rejects the \decider's waypoint in the arched doorway (red), because the corridor is blocked after 2.39\,m; the \explorer re-targets it to a reachable floor point in the same direction (green) without another API call.}
\label{fig:case265}
\end{figure}
\vfill
\begin{figure}[H]
\centering
\setlength{\tabcolsep}{2pt}
\begin{adjustbox}{max width=\linewidth}\begin{tabular}{ccc}
\small RGB: step 3, front view & \small RGB: step 4, stop & \small Map \\
\includegraphics[height=4.4cm]{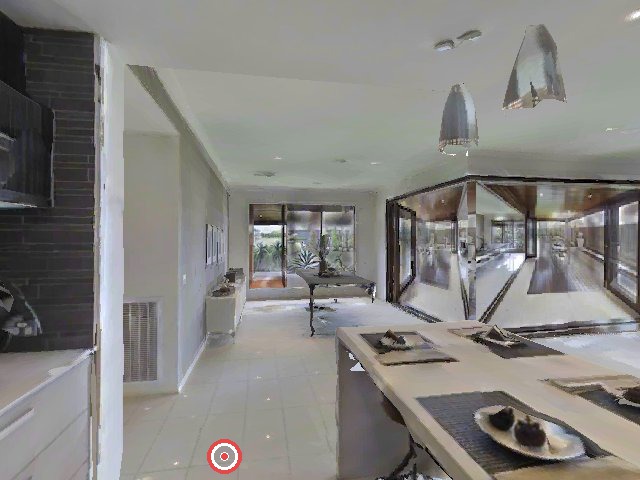} & \includegraphics[height=4.4cm]{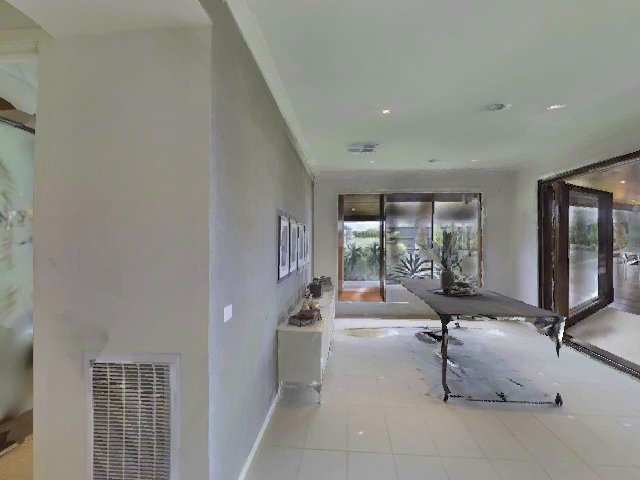} & \includegraphics[height=4.4cm]{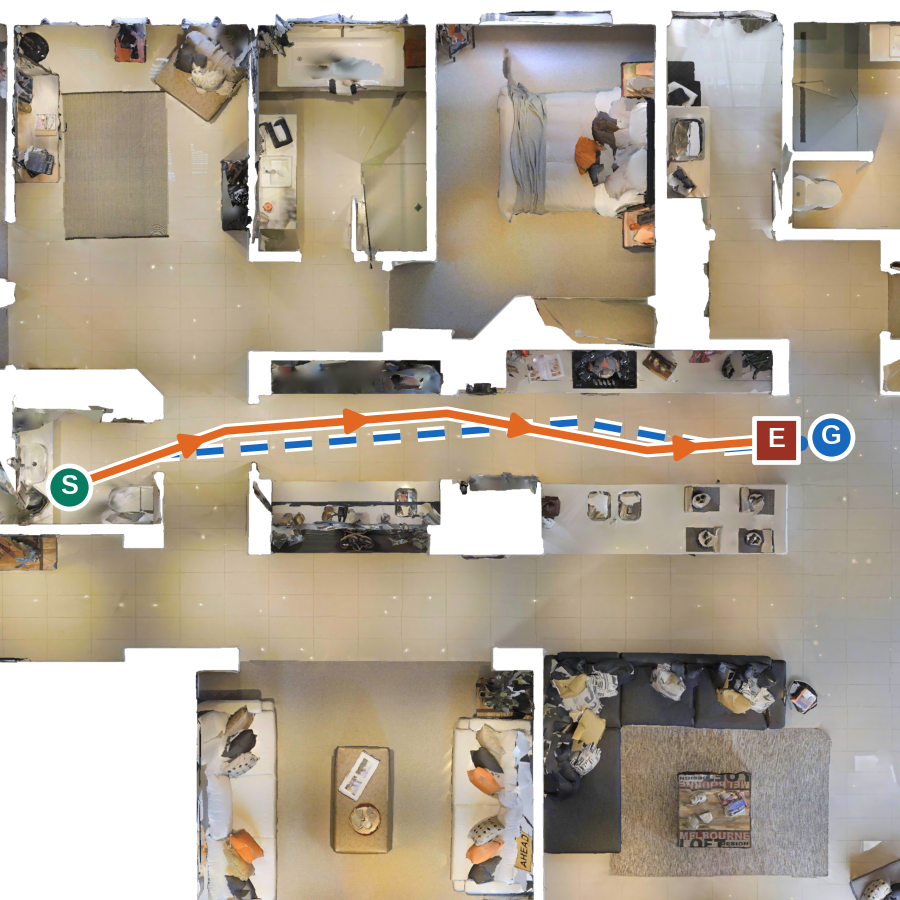} \\
\end{tabular}\end{adjustbox}
\caption{\textbf{Episode 387: the \decider continues past a premature stop.} \emph{Turn to the right and go out the bathroom door. Turn to the right a little more. Go through the galley kitchen. Walk between the stove and the island. When you get past the island, stop, and wait.} At step 3 the \explorer hands off with an intended \st before the end of the island; the \decider instead moves 1.8\,m further past the island (red) and stops at step 4, 0.74\,m from the goal (SPL 1.00).}
\label{fig:case387}
\end{figure}
\vfill
\begin{figure}[H]
\centering
\setlength{\tabcolsep}{2pt}
\begin{adjustbox}{max width=\linewidth}\begin{tabular}{ccc}
\small RGB: step 0, right view & \small RGB: step 4, stop & \small Map \\
\includegraphics[height=4.4cm]{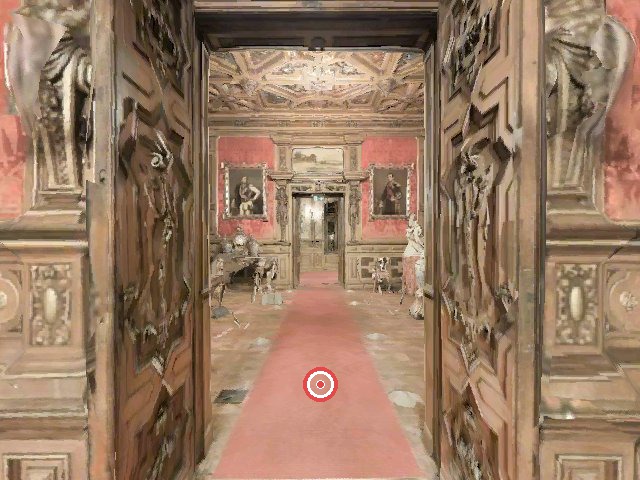} & \includegraphics[height=4.4cm]{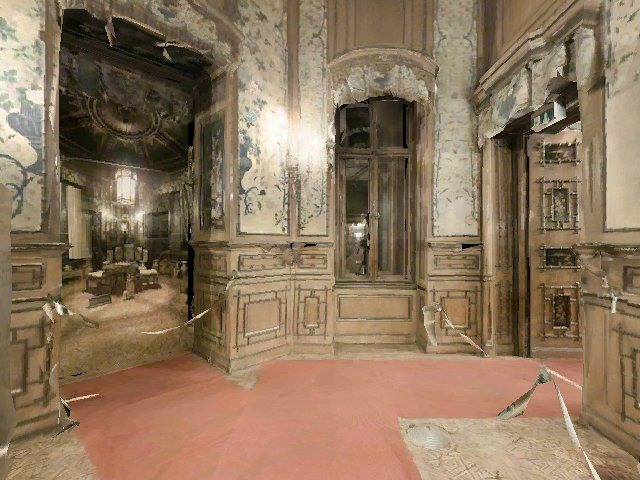} & \includegraphics[height=4.4cm]{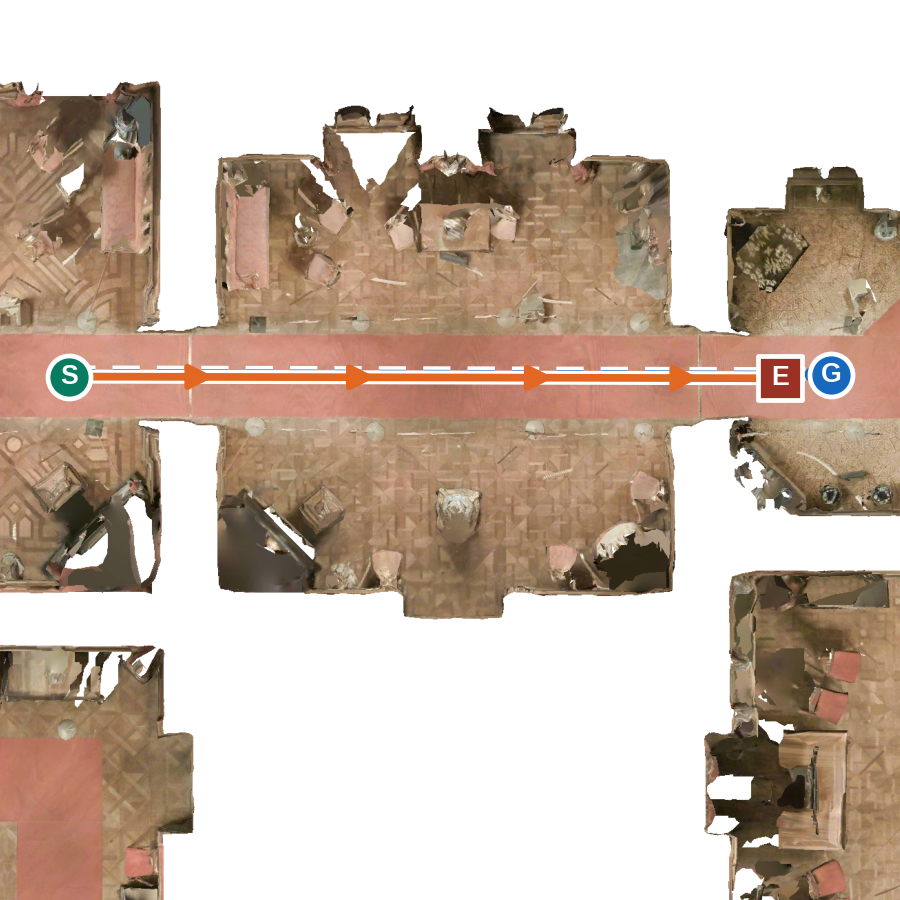} \\
\end{tabular}\end{adjustbox}
\caption{\textbf{Episode 559: following a sequence of landmarks.} \emph{Walk straight out the door in front of you and follow the red carpet. Keep going through the room with the ropes and stop when you enter the next room with ropes.} The \decider follows the carpet through the first roped room in four moves and stops on entering the next one, 0.80\,m from the goal (SPL 1.00, nDTW 0.97).}
\label{fig:case559}
\end{figure}

\clearpage
\subsection{Failure cases}

\begin{figure}[H]
\centering
\setlength{\tabcolsep}{2pt}
\begin{adjustbox}{max width=\linewidth}\begin{tabular}{ccc}
\small RGB: step 11, back view & \small RGB: step 29, stop & \small Map \\
\includegraphics[height=4.4cm]{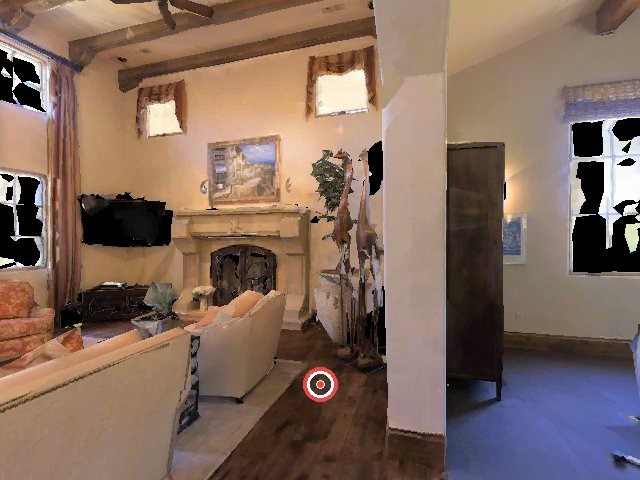} & \includegraphics[height=4.4cm]{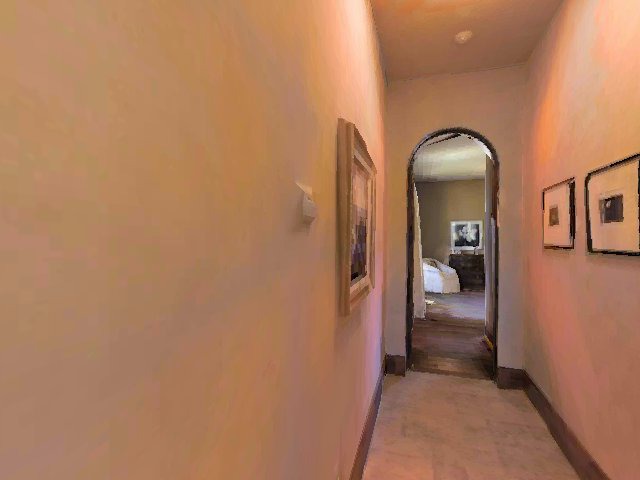} & \includegraphics[height=4.4cm]{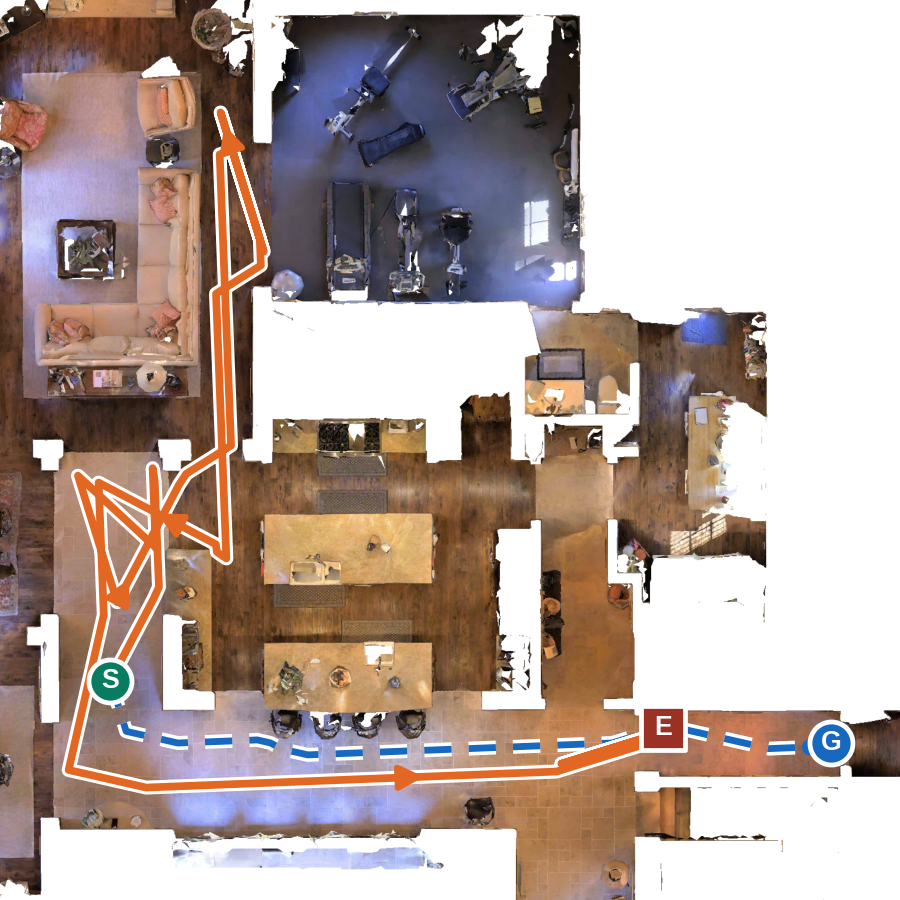} \\
\end{tabular}\end{adjustbox}
\caption{\textbf{Episode 94: oscillation after forced moves.} \emph{Walk forward, take a left around the corner and walk all the way to the end of the hallway, and stop at the door before the bedroom.} In most steps the \decider spends its information budget on recalls of past views and then moves in the forced round; between steps 11 and 14 it reverses its heading four times in a row while searching for the hallway. It reaches the end of the hallway only at step 29 and stops 3.19\,m from the goal, after a 61.6\,m path for a 13.9\,m route.}
\label{fig:case94}
\end{figure}
\vfill
\begin{figure}[H]
\centering
\setlength{\tabcolsep}{2pt}
\begin{adjustbox}{max width=\linewidth}\begin{tabular}{ccc}
\small RGB: step 0, back view & \small RGB: step 2, stop & \small Map \\
\includegraphics[height=4.4cm]{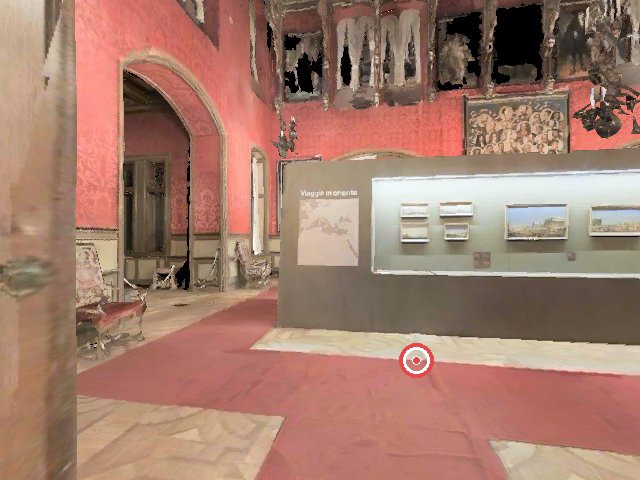} & \includegraphics[height=4.4cm]{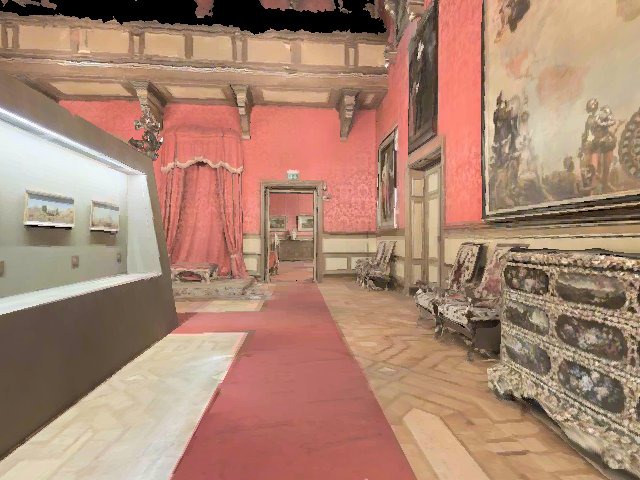} & \includegraphics[height=4.4cm]{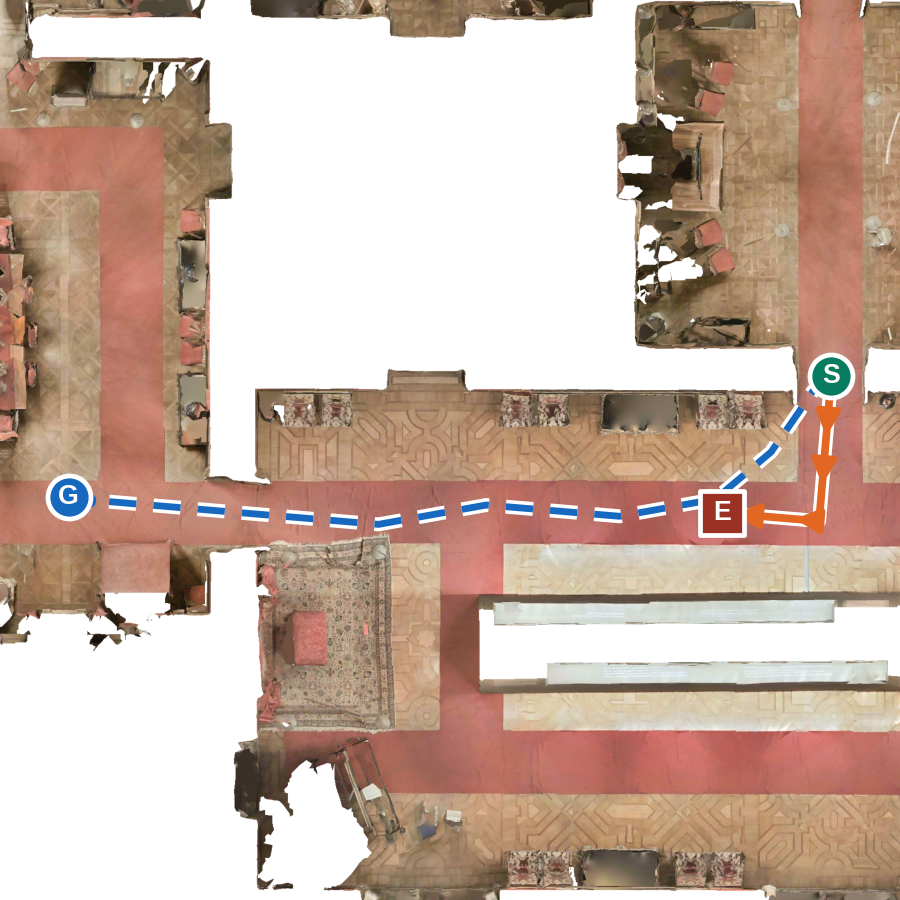} \\
\end{tabular}\end{adjustbox}
\caption{\textbf{Episode 469: stopping at the wrong landmark.} \emph{Turn right and walk along the red carpet to your right. Walk past the doorway wait by the glass info panes.} A glass display is visible behind the agent at the start; the \decider follows the carpet towards it (red) and stops beside it after two moves (5.0\,m), whereas the reference route passes the display and continues along the carpet to a goal 13.4\,m away.}
\label{fig:case469}
\end{figure}
\vfill
\begin{figure}[H]
\centering
\setlength{\tabcolsep}{2pt}
\begin{adjustbox}{max width=\linewidth}\begin{tabular}{ccc}
\small RGB: step 4, front view & \small RGB: step 5, stop & \small Map \\
\includegraphics[height=4.4cm]{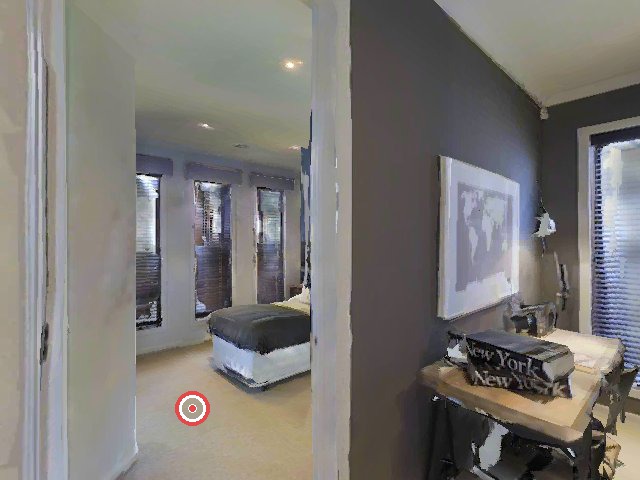} & \includegraphics[height=4.4cm]{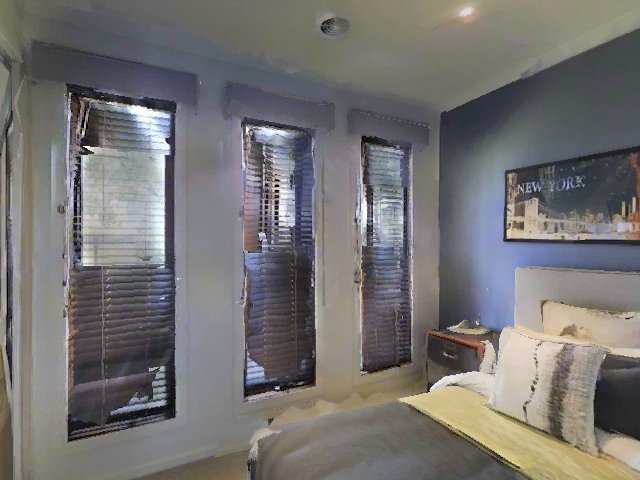} & \includegraphics[height=4.4cm]{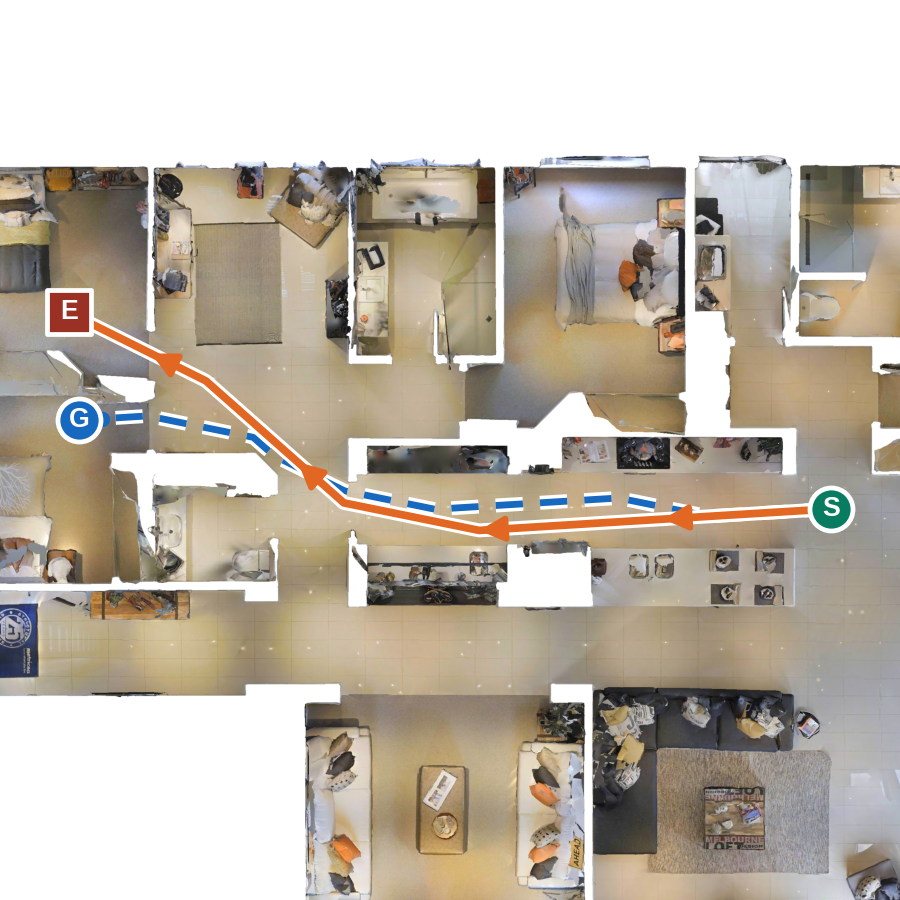} \\
\end{tabular}\end{adjustbox}
\caption{\textbf{Episode 715: entering the neighbouring room.} \emph{walk through kitchen, turn right at bathroom, immediately left again and walk through doorway to bedroom and stop.} The agent follows the reference route through the kitchen and past the bathroom (nDTW 0.86), but walks through the doorway of the neighbouring bedroom (red) and stops 3.74\,m from the goal, just outside the 3\,m success radius.}
\label{fig:case715}
\end{figure}
\endgroup
\clearpage

\section{Limitations and future work}
\label{sec:limita}

Like other training-free navigation agents built on API models, \method inherits the properties of the models it calls: API models are updated and retired over time and respond stochastically, so results obtained with one model version may not be exactly reproducible with the next.
Moreover, every decision step still involves several model calls, and their latency, together with the noise of real-world depth and odometry, remains a challenge for deployment on physical robots.
Our findings suggest two directions for future work.
First, starting from open-source models, a specialized \explorer could be trained for evidence gathering, \eg, by learning from the evidence that the \decider requests after a handoff and by equipping the \explorer with navigation skills such as landmark-directed depth measurement and selective recall of past views, which could provide more useful evidence at a lower local cost.
Second, since the remaining API cost is dominated by the visual context that the \decider receives, in particular its recalls of past views, more compact forms of evidence, such as cropped regions of past views, textual summaries of visited places, or read budgets adapted to each step, could further reduce API consumption without degrading the \decider's decisions.

